\documentclass[11pt]{article}

\usepackage[final]{acl}

\usepackage{times}
\usepackage{latexsym}

\usepackage{amsmath,amssymb}
\usepackage{booktabs}
\usepackage{array}
\usepackage{tabularx}

\usepackage[T1]{fontenc}

\usepackage[utf8]{inputenc}

\usepackage{microtype}

\IfFileExists{inconsolata.sty}{\usepackage{inconsolata}}{}

\usepackage{graphicx}

\newcommand{\TripletCorrectness}{C_{\text{triplet}}}

\title{Ontology-Guided Knowledge Graph Construction from Structured Metadata and Climate Science Publications}

\author{
 \textbf{Aayush Acharya\textsuperscript{1}},
 \textbf{Qi Zhang\textsuperscript{1}},
 \textbf{Eduard Dragut\textsuperscript{1}},
 \textbf{Longin Jan Latecki\textsuperscript{1}},
\\
 \texttt{\{aayush.acharya,qi.zhang,edragut,latecki\}@temple.edu}
\\
 \textsuperscript{1}Temple University,
\\
 \small{
   \textbf{Correspondence:} \href{mailto:aayush.acharya@temple.edu}{aayush.acharya@temple.edu}
 }
}

\begin{document}
\maketitle
\begin{abstract}
The rapid expansion of scientific literature far outpaces the speed of manual curation, creating a fundamental bottleneck in extracting structured knowledge from scientific texts. To address this, we introduce an agentic, ontology-guided framework for knowledge graph (KG) construction. 
Our framework utilizes specialized agents for document understanding, information extraction, ontology conformance, and entity linking. 
First, an LLM extracts factual knowledge in the form of triples with explicit textual evidence from research papers. 
Then the triplets undergo a multistage verification process to ground the triplets in the ontology. 
By enforcing schema constraints and requiring evidence in the form of text fragments, we ensure the data is both semantically consistent and traceable. In an agreement study over 1,000 extracted triplets, Gemini-2.5-flash achieves Cohen's $\kappa$ scores of 0.640--0.700 against three human annotators, matching the human--human agreement range of 0.616--0.689. We applied this framework to 2,000 scientific publications and construct a novel, ontology-grounded KG for climate science with 133,990 nodes and 755,308 relations; of the 23,327 triplets that the publications contribute after validation, 98.96\% are judged correct. The pipeline code and KG is publicly available\footnote{\url{https://github.com/aayushacharya/climatekg}}.
\end{abstract}

\section{Introduction}

\label{sec:Into}

The volume and complexity of the scientific literature continue to grow at an unprecedented rate \cite{hanson_strain_2024,ding2025scitoolagent}, challenging the ability of researchers to synthesize knowledge across publications, models, and domains. Knowledge graphs (KGs) provide a structured representation of scientific knowledge that can support semantic search, retrieval-augmented generation, and scientific reasoning \cite{peng_knowledge_2023,guan-etal-2025-kg}. From an NLP perspective, constructing such graphs from unstructured text requires robust information extraction (IE) systems capable of identifying entities, relations, and provenance while respecting domain semantics.

Recent progress in Large Language Models (LLMs) has significantly improved IE performance \cite{wei_chatie_2024}, enabling zero-shot and few-shot extraction of relational facts from scientific text \cite{kartchner-etal-2023-zero}. The large context window of newer LLMs allows the processing of documents without the need to distribute them into chunks. 
Moreover, the shift from simple text processing to ``native'' understanding of document layouts, tables, and visual hierarchies without needing separate OCR greatly improves IE \cite{ke_large_2026}. A parallel line of work has explored ontology-guided IE, where schemas constrain extraction to a fixed set of entity and relation types \cite{van_cauter_ontology-guided_2024}. Due to semantic consistency and interoperability, ontology-guided methods consistently show better extraction quality in domain specific setting \cite{bhattarai_document-level_2024, cotti_ontologx_2025}.

Climate science exemplifies both the opportunity and the challenge. Recent works \cite{chen_climatechat_2025, bulian_assessing_2024} note that information about climate science is underrepresented in current LLMs. Climate models form the backbone of climate research and assessment, yet critical information about model genealogy, shared components, physical parameterizations, and experimental setups is dispersed across thousands of publications \cite{kuma2023climate}.
While controlled vocabularies such as Earth System Grid Federation (ESGF), a global infrastructure that standardizes and distributes climate model datasets, and Coupled Model Intercomparison Project Phase 6 (CMIP6), an international framework defining standardized climate model experiments and metadata, exist, they primarily describe datasets and metadata rather than the rich relational knowledge embedded in the scientific literature. To date, there is very little work done on comprehensive KG integrating these vocabularies with systematically extracted, literature-derived knowledge about climate models, which is the focus of this work.

Climate science further poses a distinctive challenge for LLM-based extraction due to its reliance on domain-specific, long-tail entities and implicit dependencies that are rarely stated explicitly. For example, in the statement: 
\begin{quote}\texttt{CESM2} \emph{shares atmospheric component} with \texttt{CAM6}.\end{quote} 

an extraction system must infer that \texttt{CESM2} and \texttt{CAM6} are climate models, that the relation corresponds to an ontology-defined predicate such as \emph{hasComponent}, and that \texttt{CAM6} represents an atmospheric component rather than a standalone system. Such inferences go beyond surface-level pattern matching and expose a key limitation of open extraction and general-purpose LLMs, which often lack familiarity with specialized climate terminology and its implicit semantics. This motivates the need for ontology-guided extraction that explicitly encodes domain knowledge, constrains relation semantics, and supports reasoning over long-tail scientific entities.

In this paper, we address these gaps by introducing an evidence-based, ontology-guided IE framework and using it to construct a novel climate KG. Our approach requires each extracted subject--predicate--object triplet to be accompanied by explicit textual evidence and applies automated LLM-based validation to verify both the triplet and its evidence. This design reduces human annotation effort by narrowing manual review to concise evidence spans rather than entire documents, while maintaining high quality through automated filtering. We show that LLM-based validation achieves high agreement with human annotators, enabling scalable KG construction with minimal human intervention.
In summary, our  contributions are:
\begin{itemize}

    \item \textbf{Formulation.} We formalize domain-specific KG construction as an \emph{evidence-grounded extraction task}. By first extracting triples with explicit textual evidence, we streamline complex literature curation into a more efficient verification process for domain-specific KGs.

    \item \textbf{Framework.} We propose an agentic, ontology-guided framework that utilizes specialized LLM agents for document understanding, schema conformance, and hybrid entity linking. The system integrates authoritative climate science vocabularies and a dedicated evidence validator, thereby ensuring that all extracted relations are traceable to the source text.

    \item \textbf{Experiments.} In a 1,000-triplet validation study, Gemini-2.5-flash achieves Cohen's $\kappa$ scores of 0.640--0.700 against three human annotators, comparable to human--human agreement.
    Furthermore, 98.96\% of the 23,327 triplets that survive validation are judged correct, and a stage ablation shows where that precision comes from, at a verification cost of roughly 1\% of extraction.

     \item \textbf{Climate KG.} We introduce a novel climate KG that unifies ESGF metadata with CMIP6 controlled vocabularies, concepts from GCMD (Global Change Master Directory), a NASA-maintained controlled vocabulary that standardizes Earth science metadata to support consistent data discovery and interoperability, and GeoNames-based geolocation. It is further enriched with evidence-grounded relations extracted from 2,000 scientific publications, yielding 133,990 nodes and 755,308 relations that connect models, experiments, variables, institutions, and regions with literature-derived, text-verifiable links; 73,283 of those nodes and 298,747 of those relations come from the publication corpus rather than from the initialization sources.

\end{itemize}

\section{Problem Formulation}
\label{sec:problem_info}
Let $\mathcal{O} = (\mathcal{C}, \mathcal{R})$ denote a domain ontology, where
$\mathcal{C}$ is the set of entity classes and $\mathcal{R}$ is the set of typed relations with
explicit domain and range constraints.
Let $\mathcal{D} = \{d_1, \ldots, d_N\}$ be a corpus of scientific publications.

Our objective is to construct a domain-specific KG
$G = (V, E)$, where each edge $r \in E$ is a triple $(h, r, t)$ grounded in
explicit textual evidence.
For each document $d_i$, we aim to extract a set of evidence-backed triples
\[
\mathcal{T}_i = \{(h, r, t, \varepsilon)\},
\]
where $\varepsilon$ is a sentence or paragraph(s) in $d_i$ that explicitly supports the relation.
The KG is updated iteratively.
Given a current KG $G$, the enriched KG is obtained as
\[
G = G \cup \bigcup_{i=1}^{N} \mathcal{T}_i.
\]
The initial $G$ is extracted from 
established, trusted scientific sources, as discussed in Section ~\ref{subsec:graph_initialization}.

\begin{figure*}
    \centering
    \includegraphics[width=0.9\linewidth]{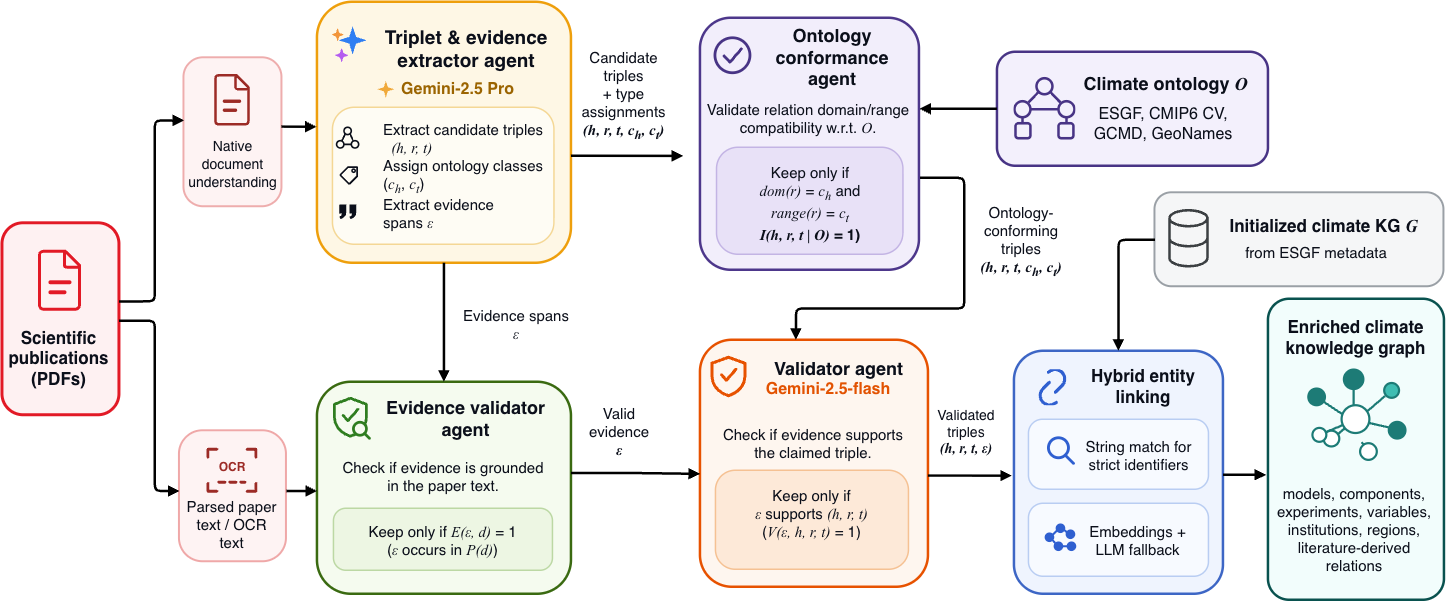}
    \caption{Ontology-guided, evidence-based KG construction pipeline. An LLM extracts candidate triples and evidence from scientific PDFs, which are validated for ontology conformance and textual grounding before final entity linking and acceptance.}
    \label{fig:info_extraction}
    \vspace{-10pt}
\end{figure*}

\section{Methodology}
\subsection{Ontology Construction}

Purely open IE suffers from uncontrolled relation phrasing and weak schema
alignment, resulting in inconsistent and difficult-to-curate outputs.
Ontology-guided extraction mitigates these issues by constraining entity types and relations to
a predefined vocabulary and by enforcing semantic consistency through domain and range validation.
Recent work has shown that incorporating ontological structure into LLM-based extraction
substantially improves relation consistency and downstream usability \cite{nie_knowledge_2024}.

We design a two-level ontology that combines standardized scholarly metadata with domain-specific concepts, using OpenAlex-derived paper-centric entities as a stable backbone.
Domain-specific classes are introduced only for screened, domain-relevant publications, ensuring scalable extraction while minimizing noise and schema drift.

\subsubsection{Domain-Specific Ontology}

The domain ontology is constructed with input from a domain expert by unifying authoritative climate resources: ESGF metadata, GCMD \cite{Nagendra2001NASAGC}, CMIP6 Controlled Vocabularies, and GeoNames. These sources provide standardized model, experiment, variable, component, institution, and geographic concepts; additional details are provided in Appendix~\ref{subsec:app_justification_ontology}.
The resulting ontology $\mathcal{O}$ defines domain and range constraints such as
$\text{dom}(\texttt{hasComponent}) = \texttt{Model}$ and also
$\text{range}(\texttt{hasComponent}) = \texttt{Component}$.
These constraints serve as soft structural priors during information extraction.

\subsection{Scientific Publication Corpus Collection}
\label{subsec:corpus}
We assemble a targeted corpus of domain-specific scientific literature to ensure coverage of authoritative and highly relevant sources. The first phase begins with a small seed set of core papers and expands through two levels of citation mining with LLM-based relevance filtering. After the first phase selected 1,007 papers, we used the top OpenAlex concepts from this set to retrieve recent, context-matched candidates and applied the same screening procedure, adding 993 papers for a 2,000-paper corpus; further details are in Appendix~\ref{subsec:app_screening}.
The small seed of core papers for climate-domain task was taken from the following sources. The sources were selected on consultation with climate experts:
(1) curated references from the IPCC Sixth Assessment Report \cite{intergovernmental_panel_on_climate_change_ipcc_annex_2023} and
(2) CMIP6 publications\footnote[1]{https://cmip-publications.llnl.gov/}. 

\subsection{Ontology-Guided IE Pipeline}
\label{subsec:ie_pipeline}

We propose an ontology-guided IE pipeline that integrates semantic constraints
and explicit evidence grounding throughout the extraction process.
The agentic system is presented in Figure ~\ref{fig:info_extraction}.
Now we give a brief overview of the role of each agent represented by an LLM.

\paragraph{Triplet and evidence extractor agent}
The detailed architecture for information extraction pipeline is given in Figure~\ref{fig:info_extraction}. For each document $d\in D$, we predict triplets $(h_1, r_1, t_1), \dots,(h_N,r_N,t_N)$ with evidence spans $\epsilon_1 \ldots \epsilon_n$ supporting each triplet. For each triplet $(h,r,t)$, we also predict its ontology classes $c_h$ and $c_t$.

\noindent\textit{Example.}
From a single sentence the agent may predict
\[
\begin{aligned}
(h_1, r_1, t_2) = (&\text{CESM2},\ \texttt{hasComponent},\\
&\text{CAM6}),
\end{aligned}
\]
\[
\begin{aligned}
(h_2, r_2, t_2) = (&\text{CAM6},\ \texttt{appliesToRealm},\\
&\text{atmos})
\end{aligned}
\]
together with ontology-aligned type assignments
\[
c_{h1} = \texttt{Source}, \quad c_{t1} = \texttt{SourceComponent},
\]
\[
c_{h2} = \texttt{SourceComponent}, \quad c_{t2} = \texttt{Realm}
\]
and an explicit supporting evidence span

\noindent
$\epsilon_1,\epsilon_2 =$ \emph{CESM2 uses CAM6 as its atmospheric component to simulate large-scale atmospheric circulation.}

The agent uses Gemini's native document understanding to extract these features. Beyond the triplets, there are other features extracted during this stage which is provided in Appendix \ref{subsec:document_level_features}.

\paragraph{Ontology conformance agent}
Candidate triples are first checked for ontology conformance:
\[
\mathbb{I}(h,r,t \mid \mathcal{O}) =
\textbf{1}[\text{dom}(r)=c_h \land \text{range}(r)=c_t].
\]
Only if $ \mathbb{I}(h,r,t \mid \mathcal{O}) = 1$,
a triplet is passed to the evidence validator agent. The check is deterministic and is specified in Appendix~\ref{subsec:app_conformance_spec}. 

\noindent\textit{Example.}
Assume the ontology $\mathcal{O}$ defines the relation $\texttt{hasComponent}:$ $\texttt{Source} \rightarrow \texttt{SourceComponent}$. Consider the following sentence in document $d$:
\begin{quote}
CAM6 is the atmospheric component used within the CESM2 climate modeling framework.
\end{quote}
From this sentence, an LLM-based extractor may plausibly generate the triple:
\[
(h, r, t) = (\text{CAM6},\ \texttt{hasComponent},\ \text{CESM2}),
\]
by incorrectly assigning the syntactic subject as the head entity. The inferred ontology classes are:
\[
c_h = \texttt{SourceComponent}, \quad c_t = \texttt{Source}.
\]

However, this assignment violates the ontology constraints for relation $\texttt{hasComponent}$. As a result, the ontology conformance agent returns:
\[
\mathbb{I}(h,r,t \mid \mathcal{O}) = 0,
\]
and the triplet is discarded before any evidence or semantic check is run on it.

\paragraph{Evidence validator agent}
Unlike other multi-agent information extraction frameworks \cite{lu_karma_2025}, which enforce ontology-level constraints without explicitly verifying textual grounding, our framework introduces a dedicated evidence validator agent. It ensures that the extracted sentence/paragraph is not introduced through hallucination or paraphrasing.
Given a candidate evidence span $\varepsilon$ and the parsed content of the paper
$\mathcal{P}(d)$, the agent verifies whether $\varepsilon$ occurs verbatim or as a faithful
substring within the document. 
Formally, we define an evidence presence function
\[
\mathbb{E}(\varepsilon, d) =
\begin{cases}
1, & \text{if } \varepsilon \subseteq \mathcal{P}(d), \\
0, & \text{otherwise}.
\end{cases}
\]
Only triples $(h,r,t,\varepsilon)$ satisfying $\mathbb{E}(\varepsilon,d)=1$ are forwarded to
the validator agent, so that every triple reaching semantic validation carries an evidence span that
is present in the source document.
Triples with missing or hallucinated evidence are discarded at this point.
Conformance and evidence grounding are independent tests, so applying them in either order yields the
same surviving set; we run conformance first because it is the cheaper of the two and removes the
largest share of candidates.

\paragraph{Validator agent}
It receives the triples that pass both the ontology conformance agent and the evidence validator agent.
It validates whether each triplet $ (h, r, t)$ of classes $c_h$ and $c_t$ is indeed supported by  evidence $\varepsilon$:
\[
\mathbb{V}(\varepsilon, h, r, t) =
\begin{cases}
1, & \text{if } \varepsilon \text{ supports } (h,r,t), \\
0, & \text{otherwise}.
\end{cases}
\]
Triples with $\mathbb{V}(\varepsilon, h, r, t)=0$ are discarded.

\subsection{Graph Initialization}
\label{subsec:graph_initialization}

We initialize the KG $G$ from over four million ESGF dataset metadata records\footnote{Retrieved from the LLNL ESGF index through the \texttt{esg-search} Solr API (\url{https://esgf-node.llnl.gov/esg-search/search}, queried through the AIMS2 proxy at \url{https://aims2.llnl.gov/proxy/search}) in a snapshot taken between April and May 2025. The CMIP6 controlled vocabularies are the WCRP \texttt{CMIP6\_CVs} release (\url{https://github.com/WCRP-CMIP/CMIP6_CVs}) of the same period.}, deterministically mapping standardized model, experiment, variable, institution, and spatial attributes into ontology-aligned entities and relations. This trusted backbone grounds subsequent literature-derived enrichment by giving the entity linker canonical climate identifiers instead of relying on the LLM to generate or memorize standardized names. Additional construction details are provided in Appendix~\ref{subsec:app_graph_initialization}.

\begin{figure*}
    \centering
    \includegraphics[width=0.7\linewidth]{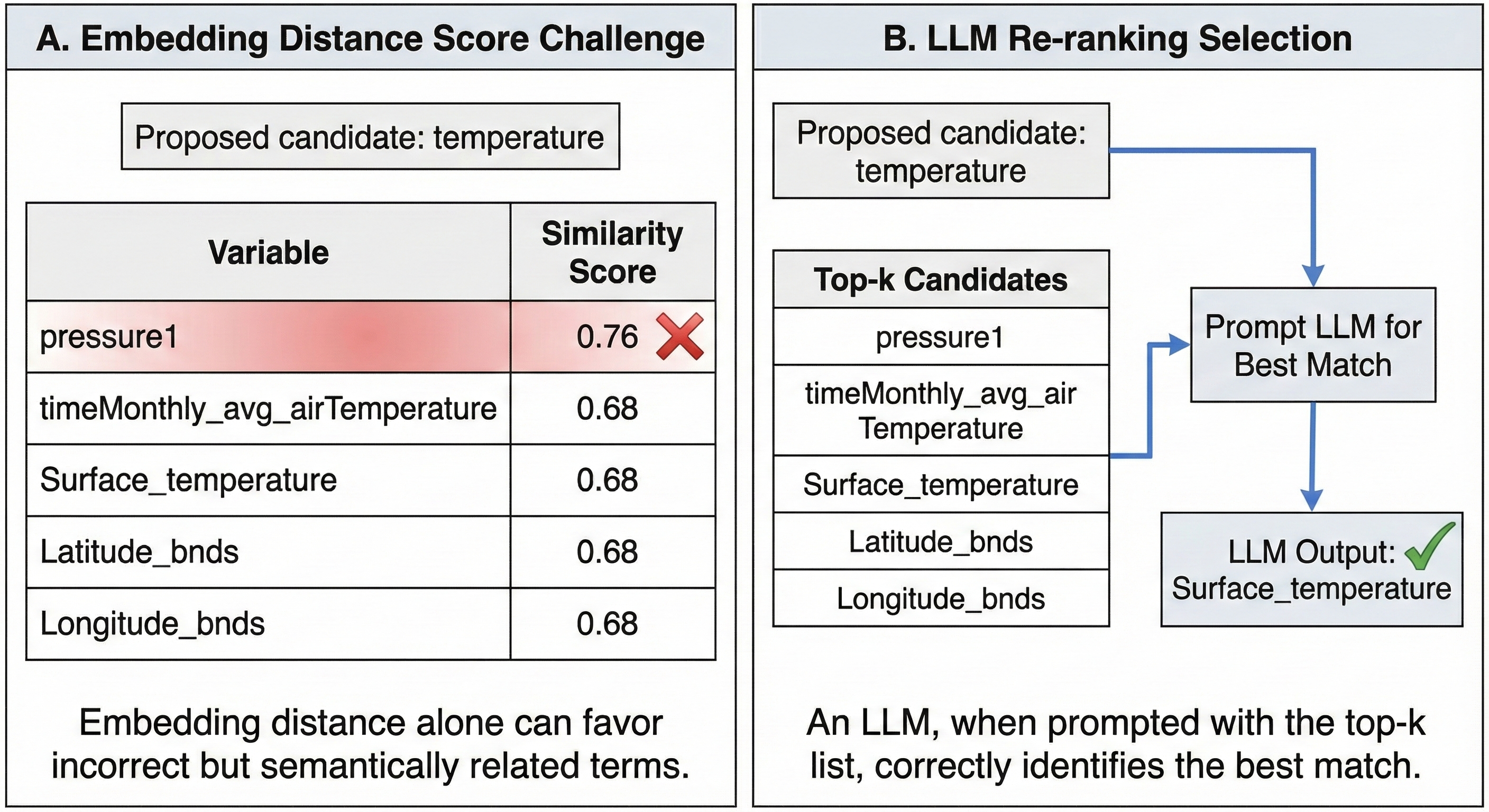}
    \caption{Hybrid entity linking pipeline showing how embedding distance fails in cases where the similarity score is low and LLM selection helps.}
    \label{fig:hybrid_entity_linking}
\end{figure*}

\subsection{Hybrid Entity Linking}

This section describes the Entity Linking agent (Figure~\ref{fig:entity_linking}, Appendix).
For each triple $(h,r,t)$ and evidence $\epsilon$, we consider $(h,t)$ as candidate entities to be linked to existing nodes of $G=(V,E)$.
For linking, we consider string-based similarity or embedding-based similarity, depending on the type of entity. 

We use 
NV-Embed-V2 \cite{lee_nv-embed_2025} 
to produce deep embeddings, since it
showed consistently good performance in MTEB benchmark \cite{muennighoff2022mteb,enevoldsen2025mmtebmassivemultilingualtext} and was also utilized by \cite{pan_taxonomy-driven_2025} for Climate tasks. In general, the embedding model performs well, but the embedding distance is widely inconsistent for similar items. 
Given a deep embedding function $\psi$,
let $\hat{v}$ be the most similar graph node to an entity $e$:
\[
\hat{v} = \arg\max_{v \in V} \text{sim}(\psi(e), \psi(v)),
\]
If $\text{sim}(\psi(e), \psi(\hat{v})) \geq \tau$, $\hat{v}$ is linked to $e$. However, if $\text{sim}(\psi(e), \psi(\hat{v})) < \tau$, we retrieve the top-$k$ candidates ${v_1,\ldots,v_k}$ using a fallback threshold $\tau_f$ and ask an LLM to identify the best match given the provided evidence.
We use LLM-based candidate selection, as shown in Figure~\ref{fig:hybrid_entity_linking}, because embedding similarity can be unreliable for specialized climate terms, where distributional proximity does not always reflect ontological compatibility. Each design choice in the pipeline is settled empirically rather than by preference: the linking thresholds come from the sensitivity analysis of Appendix~\ref{subsec:app_entity_linking_threshold_sensitivity} (Table~\ref{tab:entity_linking_threshold_sensitivity}), the composition of verification stages from the ablation of Table~\ref{tab:stage_ablation}, and the choice of a deterministic test over an LLM call for ontology conformance and evidence grounding from the cost comparison in Appendix~\ref{subsec:app_extraction_cost_runtime}.
We empirically set the automatic linking threshold $\tau = 0.80$ and the LLM fallback threshold $\tau_f = 0.55$. These thresholds are selected using the sensitivity analysis(refer to Appendix, Table~\ref{tab:entity_linking_threshold_sensitivity}), where match rates rise sharply above 0.80 and bins down to 0.55 still recover useful LLM-confirmed matches. For strict identifiers such as climate model names, we use string matching rather than embedding similarity; details are provided in Appendix~\ref{subsec:app_linking_strategy}.

It is possible that some facts are repeated by multiple papers during this stage. The handling is provided in Appendix \ref{subsec:repeated_facts}.

\begin{table*}[!t]
\centering
\scriptsize
\setlength{\tabcolsep}{4pt}
\caption{Knowledge saturation over the 2,000-publication entity-linking run. Novelty is computed within each paper interval as the proportion of unique items that had not appeared in earlier papers. Relation novelty is computed over unique source--relation--target assertions.}
\label{tab:knowledge_saturation}
\begin{tabular*}{0.98\textwidth}{@{\extracolsep{\fill}}lccccc@{}}
\toprule
\textbf{Paper Interval} & \textbf{New Entities / Paper} & \textbf{Entity Novelty (\%)} & \textbf{New Mentions / Paper} & \textbf{Mention Novelty (\%)} & \textbf{Relation Novelty (\%)} \\
\midrule
1--500 & 16.89 & 61.9 & 19.88 & 43.3 & 84.4 \\
501--1{,}000 & 12.31 & 43.7 & 13.98 & 26.9 & 77.0 \\
1{,}001--1{,}500 & 10.58 & 46.1 & 12.40 & 28.9 & 77.5 \\
1{,}501--2{,}000 & 8.81 & 40.3 & 10.10 & 24.1 & 73.1 \\
\bottomrule
\end{tabular*}
\end{table*}

\begin{figure*}[!t]
    \centering
    \includegraphics[width=0.95\linewidth]{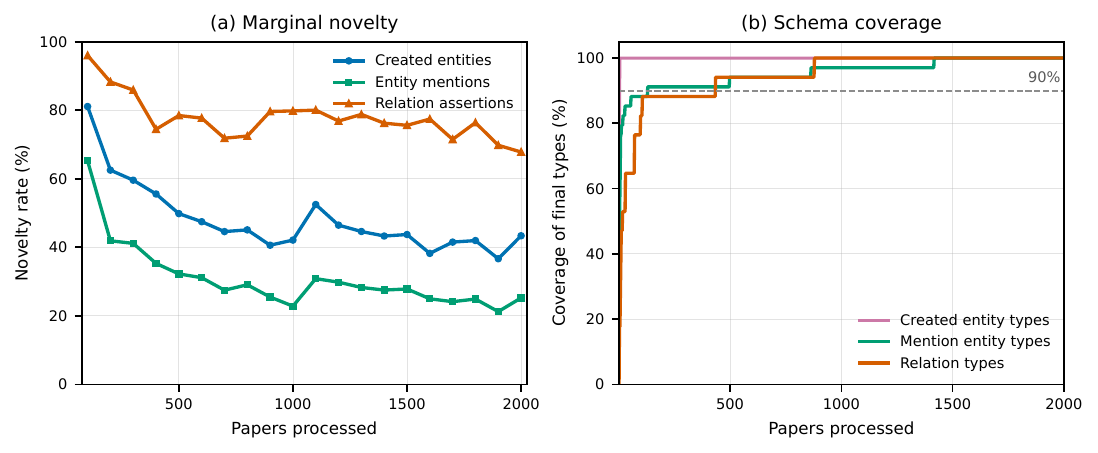}
    \caption{Knowledge saturation as papers are added to the KG. Left: novelty rates are computed over non-overlapping 100-paper windows. Right: schema coverage is the cumulative fraction of final entity or relation types observed after each paper.}
    \label{fig:knowledge_saturation}
\end{figure*}

\section{Experimental Evaluation}

\subsection{LLM and Human Annotator Agreement}
\label{subsec:annotator_agreement}
To assess whether LLM-based validation can scale KG quality control, we compare Gemini-2.5-flash with three human annotators on 1,000 extracted triplets sampled from IPCC-linked climate-science papers \cite{intergovernmental_panel_on_climate_change_ipcc_annex_2023} of the corpus of Section~\ref{subsec:corpus}. The sample is drawn from the raw output of the triplet and evidence extractor agent, before any verification stage, so it contains triplets that the pipeline later keeps as well as triplets it removes. Each decision is the binary judgement of whether the evidence span supports the triplet, which is the task of the validator agent in Figure~\ref{fig:info_extraction}. As shown in Table~\ref{tab:annotator_agreement}, Gemini-human Cohen's $\kappa$ scores range from 0.640--0.700, comparable to the human-human range of 0.616--0.689; against the human majority vote, Gemini reaches $\kappa=0.798$ and 0.977 observed agreement. This supports its use as a scalable surrogate validator for large-scale KG construction.

\begin{table}[!t]
\centering
\small
\caption{Agreement among human annotators and Gemini-2.5-flash on 1,000 triplet validation decisions.}
\label{tab:annotator_agreement}
\begin{tabular}{lcc}
\toprule
\textbf{Comparison} & \textbf{Obs. Agreement} & \textbf{Cohen's $\kappa$}  \\
\midrule
Ann. 1 vs. Ann. 2 & 0.946 & 0.616 \\
Ann. 1 vs. Ann. 3 & 0.956 & 0.667 \\
Ann. 2 vs. Ann. 3 & 0.960 & 0.689 \\
Gemini vs. Ann. 1 & 0.950 & 0.640  \\
Gemini vs. Ann. 2 & 0.952 & 0.645  \\
Gemini vs. Ann. 3 & 0.962 & 0.700  \\
Gemini vs. \\ human majority & 0.977 & 0.798 \\
\bottomrule
\end{tabular}
\end{table}

\begin{table*}[!t]
\centering
\caption{Per-stage attrition over the full 2,000-publication corpus, evaluated by Gemini-2.5-flash. Each row reports the output of the cumulative pipeline through that stage.}
\label{tab:stage_attrition}
\begin{tabular}{lcccc}
\toprule
\textbf{Stage} & \textbf{Input Triplets} & \textbf{Filtered} & \textbf{Output Triplets} & \textbf{$\TripletCorrectness$ (\%)} \\
\midrule
Triplet and evidence extractor agent & -- & -- & 39{,}964 & 94.05 \\
Ontology conformance agent & 39{,}964 & 11{,}050 & 28{,}914 & 94.29 \\
Evidence validator agent & 28{,}914 & 1{,}014 & 27{,}900 &  94.72 \\
Validator agent & 27{,}900 & 4{,}573 & 23{,}327 & 98.96 \\
\bottomrule
\end{tabular}
\end{table*}

\begin{table}[!t]
\centering
\small
\caption{Stage ablation over the same 39,964 candidate triplets. O = ontology conformance agent, E = evidence validator agent, V = validator agent. Every triplet stores the verdict of every stage, so all eight variants are evaluated post hoc with the same $\TripletCorrectness$ judge.}
\label{tab:stage_ablation}
\begin{tabular}{lrr}
\toprule
\textbf{Stages applied} & \textbf{Triplets kept} & \textbf{$\TripletCorrectness$ (\%)} \\
\midrule
none (extractor) & 39{,}964 & 94.05 \\
O & 28{,}914 & 94.29 \\
E & 38{,}386 & 94.38 \\
V & 32{,}436 & 98.86 \\
O+E & 27{,}900 & 94.72 \\
O+V & 24{,}172 & 98.96 \\
E+V & 31{,}095 & 98.88 \\
\textbf{O+E+V (full)} & \textbf{23{,}327} & \textbf{98.96} \\
\bottomrule
\end{tabular}
\end{table}

\begin{table}[!t]
\centering
\small
\caption{Aggregate statistics of the constructed climate KG after enrichment with 2,000 scientific publications. The split rows separate what the deterministic initialization of Section~\ref{subsec:graph_initialization} contributes from what the extraction pipeline adds, and are computed from the graph build files, which hold two relations more than the loaded database.}
\label{tab:kg_aggregate_stats}
\begin{tabular}{lr}
\toprule
\textbf{Statistic} & \textbf{Value} \\
\midrule
Scientific publications & 2{,}000 \\
KG nodes & 133{,}990 \\
\quad from initialization & 60{,}707 \\
\quad added from publications & 73{,}283 \\
KG relations & 755{,}308 \\
\quad from initialization & 456{,}563 \\
\quad added from publications & 298{,}747 \\
Node types & 55 \\
Relation types & 59 \\
Validated publication triplets & 23{,}327 \\
\bottomrule
\end{tabular}
\end{table}

\begin{table}[!t]
\centering
\small
\caption{Comparison with KARMA over the first 1,007 scientific publications after final validation}
\label{tab:comparison}
\begin{tabular*}{\columnwidth}{@{\extracolsep{\fill}}lrr@{}}
\toprule
\textbf{Metric} & \textbf{KARMA\textsuperscript{$\dagger$}} & \textbf{Ours} \\
 & \cite{lu_karma_2025} & \textbf{(after validator)} \\
\midrule
Output Triplets & 3{,}820 & 17{,}141 \\
Valid & 3{,}525 & 16{,}981 \\
Invalid & 295 & 160 \\
$\TripletCorrectness$ (\%) & 92.277 & \textbf{99.067} \\
\bottomrule
\end{tabular*}
\par\vspace{2pt}
\begin{minipage}{\columnwidth}
\footnotesize \textsuperscript{$\dagger$}Prompts changed for the climate domain.
\end{minipage}
\end{table}

\begin{table}[!t]
\centering
\small
\caption{Comparison with ClimateIE over the 25 papers in the ClimateIE human corpus, using the ClimateIE ontology for both systems and Gemini-2.5-flash validation.}
\label{tab:climateie_comparison}
\setlength{\tabcolsep}{4pt}
\begin{tabular*}{\columnwidth}{@{\extracolsep{\fill}}lrr@{}}
\toprule
\textbf{Metric} & \textbf{ClimateIE} & \textbf{Ours (with} \\
 & \cite{pan_climateie_2025} & \textbf{ClimateIE ontology)} \\
\midrule
Output Triplets & 5{,}729 & 637 \\
Valid & 4{,}297 & 592 \\
Invalid & 1{,}432 & 45 \\
$\TripletCorrectness$ (\%) & 75.004 & \textbf{92.936} \\
\bottomrule
\end{tabular*}
\end{table}

\subsection{Evaluation metric}

We evaluate extraction quality using a triplet precision, which we call \emph{triplet correctness} metric and denote as $C_{\text{triplet}}$. It measures the proportion of triplets generated by the system that are judged to be correct by an evaluator.

Let $T$ denote the set of triplets output by a given extraction stage, and let $T_{\text{valid}} \subseteq T$ denote the subset of those triplets validated as correct by the evaluator.
 A triplet is considered correct if the relation between the source and target entities is supported by the corresponding paragraph in the source document, allowing for implicit references and anaphora resolution. The triplet correctness score is defined as:
\[
\TripletCorrectness = \frac{|T_{\text{valid}}|}{|T|}
\]

The metric is precision-like in nature, as it evaluates correctness over generated triplets without assuming an exhaustively annotated gold standard. This makes $\TripletCorrectness$ well suited to the evaluation of ontology-guided information extraction from scientific publications, where the emphasis is on producing high-quality, curator-ready KG assertions rather than maximizing coverage.

\subsection{Results and Discussion}

Table~\ref{tab:stage_attrition} shows the effect of each validation stage over 2,000 publications. The extractor produces 39,964 candidate triplets; ontology conformance, evidence validation, and final validation reduce this to 23,327 high-quality triplets with a final correctness score of 98.963\%. The resulting KG contains 133,990 nodes and 755,308 relations across 55 node types and 59 relation types (Table~\ref{tab:kg_aggregate_stats}). Separating the two sources of content, the deterministic initialization from ESGF, CMIP6 controlled vocabularies, GCMD, and GeoNames accounts for 60,707 nodes and 456,563 relations, while the publication corpus contributes 73,283 nodes and 298,747 relations, including 96,100 paper-to-entity mention edges and 6,530 distinct entity-to-entity relations asserted by the papers themselves.

\paragraph{Stage ablation.}
Because every candidate triplet stores the verdict of every stage, we can evaluate all eight combinations of the three verification stages on the same 39,964 triplets with the same $\TripletCorrectness$ judge, without re-running any agent (Table~\ref{tab:stage_ablation}). The stages act on different axes rather than on a single quality dimension. The validator agent is the precision component: removing it from the full pipeline leaves 27,900 triplets at 94.72\%, so it contributes $+4.24$ points of $\TripletCorrectness$. Ontology conformance contributes almost no precision, but that is not its function: 93.4\% of the 11,050 triplets it removes are judged textually correct, yet their (source type, relation, target type) signature lies outside the ontology, so they cannot be typed into the KG at all. Its role is schema compatibility rather than error correction. Evidence validation likewise leaves final precision unchanged (24,172 at 98.96\% without it), and instead supplies the traceability guarantee: without it, 845 triplets whose evidence span cannot be matched against the parsed source enter the KG. Precision, schema compatibility, and traceability therefore have to be read as three separate outcomes of the pipeline. The details about what the pipeline filters remove is given in Appendix \ref{subsec:filters_remove}. The verification stack is also inexpensive relative to what it corrects.  Appendix~\ref{subsec:app_extraction_cost_runtime} reports cost, runtime, and ablation effect per stage.

\paragraph{Knowledge Saturation}
We further measure knowledge saturation as papers are incrementally linked into the KG. Here, created entities are new KG nodes introduced during entity linking, while mentions are paper-level entity references resolved against the KG. Novelty is computed within each paper interval as the number of unique items not seen in earlier papers divided by the total number of unique items observed in that interval. As shown in Table~\ref{tab:knowledge_saturation} and Figure~\ref{fig:knowledge_saturation}, the average number of new unique created entities per paper drops from 16.89 in the first 500 papers to 8.81 in the final 500 papers, while the entity novelty rate drops from 61.9\% to 40.3\%. Entity mention novelty shows a similar decline, from 43.3\% to 24.1\%, and relation assertion novelty declines from 84.4\% to 73.1\%. At the schema level, all 9 created entity types are observed by paper 4, 90\% of the final 34 mention entity types are observed by paper 129, and 90\% of the final 17 relation types are observed by paper 434. These trends indicate that additional papers increasingly reuse the existing climate KG vocabulary and schema, while continuing to add validated assertions. The same redundancy is visible at the level of individual assertions: of the 6,530 distinct literature-derived edges, 11.0\% are asserted independently by two or more papers, and 1.6\% by five or more. Facts missed in one paper are therefore often recoverable from overlapping evidence in later papers, which partially mitigates per-document recall limitations at corpus scale.

\paragraph{Comparison with baselines}
Table~\ref{tab:comparison} compares our framework with KARMA~\cite{lu_karma_2025}, the closest available agentic baseline because it is also a multi-agent framework for extracting KG-style assertions from scientific publications. We are not aware of an existing agentic KG construction system designed specifically for climate-science papers, so we adapt KARMA's prompts to the climate domain and evaluate both systems on the first 1,007 publications after final validation. Our system extracts 17,141 triplets with 99.067\% correctness, while KARMA extracts 3,820 triplets with 92.277\% correctness, showing that explicit evidence grounding and ontology-conformance validation improve precision while producing a much larger validated KG.

We also compare against ClimateIE~\cite{pan_climateie_2025}, a climate-specific IE benchmark rather than an agentic KG-construction system. For this experiment, we run our pipeline on the 25-paper ClimateIE human corpus using the ClimateIE ontology for both systems, and evaluate source--relation--target triples with the same Gemini-2.5-flash validator. As shown in Table~\ref{tab:climateie_comparison}, our system produces fewer triplets but reaches substantially higher correctness: 92.936\% compared with 75.004\%. This indicates that the proposed evidence-centered agents trade some raw extraction volume for much cleaner curator-ready assertions, outperforming both the general agentic baseline and the climate-specific IE baseline in precision. 

The 1,000 human-annotated triplets of Section~\ref{subsec:annotator_agreement} support the same conclusions with human labels, since they were sampled from raw extractor output and therefore cover triplets at every filter outcome. Against the majority vote of three annotators, triplets kept by the full pipeline are 99.0\% correct (625 of 631), which closely tracks the 98.96\% judged by the LLM evaluator. 

\subsection{Recall and Coverage}
\label{subsec:recall}

$\TripletCorrectness$ is precision-like, so we complement it in two ways: we measure entity-level recall against external gold that we did not produce, and we decompose what the verification stages remove.

\paragraph{Recall against external gold.}
The CMIP Publication Hub\footnote{\url{https://cmip-publications.llnl.gov/}} hosts publication metadata submitted through a form by the papers' own authors, who select projects, activities, experiments, frequencies, sources, realms, and variables from controlled checkbox lists. This gives an independently curated annotation that is not derived from our pipeline. We scraped the hub for DOIs, queried OpenAlex for open-access PDFs, and retained the 139 papers that both downloaded and carried entity metadata. Because a curator can tag a paper with an entity that is never written in the paper (a related model family or a nearby variant, for instance), we first check whether each gold entity occurs in the parsed papers. Only 50.78\% of the curated gold entities appear in the paper text at all, which is itself evidence that exhaustive per-document gold is ill-posed in this domain. Over the entities that are recoverable from the text, our pipeline reaches 90.29\% micro-averaged recall, as shown in Table~\ref{tab:cmip_recall}.

\begin{table}[!t]
\centering
\small
\caption{Entity-level recall against CMIP Publication Hub metadata for 139 papers, computed over gold entities whose surface form is recoverable from the parsed paper text.}
\label{tab:cmip_recall}
\begin{tabular}{lrrr}
\toprule
\textbf{Entity type} & \textbf{Gold} & \textbf{Extracted} & \textbf{Recall (\%)} \\
\midrule
Source & 797 & 774 & 97.11 \\
Experiment & 196 & 162 & 82.65 \\
Frequency & 85 & 57 & 67.06 \\
Activity & 55 & 30 & 54.55 \\
\midrule
\textbf{Micro avg.} & \textbf{1{,}133} & \textbf{1{,}023} & \textbf{90.29} \\
\bottomrule
\end{tabular}
\end{table}

\section{Related work}

Automatic KG construction from text is commonly treated as a pipeline of entity extraction, relation extraction, linking, and refinement, with persistent challenges around schema conformance, scalability, and factual reliability \cite{zhong_comprehensive_2023}. Recent LLM-based systems address these challenges by decomposing KG construction into extraction, schema alignment, canonicalization, and validation stages \cite{zhang_extract_2024}. 

Ontology-guided benchmarks further emphasize that extracted facts must be evaluated not only for correctness but also for conformance to a target schema \cite{mihindukulasooriya_text2kgbench_2023}. SciER shows that entity and relation extraction over full-text scientific papers remains hard for both supervised models and LLMs, even for generic dataset, method, and task entities \cite{zhang-etal-2024-scier}.  ClimateIE provides a climate-specific benchmark for NER, relation extraction, and entity linking \cite{pan_climateie_2025}, and taxonomy-driven climate IE work shows that grounding extraction in curated vocabularies can reduce hallucination and improve precision \cite{pan_taxonomy-driven_2025}. KARMA \cite{lu_karma_2025} uses multiple agents for scientific KG enrichment, but it is not designed for a climate-specific system with domain ontology grounding, evidence-level validation, and entity linking into a curated climate KG. Our work builds on this line of research by combining ontology-constrained extraction, explicit evidence grounding, hybrid entity linking, and scalable validation to construct a large climate-science KG from scientific publications.

\section{Conclusion}

We presented an evidence-based, ontology-guided framework for constructing climate-science KGs from scientific publications. By requiring explicit evidence, validating ontology conformance, and linking entities to a trusted climate KG backbone, the framework produces high-precision, traceable relations at scale. Across 2,000 publications, the pipeline constructs a KG with 133,990 nodes, 755,308 relations, while maintaining high triplet correctness after validation.

The results show that agentic extraction benefits from domain structure: ontology constraints reduce schema drift, evidence validation filters unsupported assertions, and hybrid entity linking connects literature-derived facts to canonical climate entities. Our agreement study supports Gemini-2.5-flash as a scalable validator, and comparisons with KARMA and ClimateIE show that the proposed framework achieves substantially cleaner curator-ready assertions than both a general multi-agent scientific IE baseline and a climate-specific IE baseline. These findings demonstrate the practicality of evidence-centered extraction for constructing auditable, domain-specific scientific KGs.

\section*{Acknowledgement}
 We thank Ryan Berkheimer from NOAA National Climatic Data Center, and Paul Smith from WCRP Coupled Model Intercomparison Project (CMIP), for countless discussions and valuable feedback. We also thank Adamu Mustapha for his valuable contribution as our climate expert. We further thank Huitong Pan, Yasmin Mohammadi, and Sanobar Rustamova for their valuable contributions to our project. This work was supported by the U.S.  National Science Foundation awards III-2107213 and ITE-2333789.

\section*{Limitations}

Our pipeline performs evidence grounding, ontology checks, entity linking, and validation in addition to extraction. Two of these verification stages are deterministic and the third is a short call to a small model, so the added cost is about 1\% of extraction (Appendix~\ref{subsec:app_extraction_cost_runtime}); the pipeline is nevertheless more expensive than direct extraction, and it inherits the GPU cost of local parsing and screening. We do not compute triplet-level recall, since open-ended scientific text would require exhaustive gold triplets for each document. Instead we report entity-level recall against external curated gold restricted to what the text actually states (90.29\% micro-average, Section~\ref{subsec:recall}), and we quantify how many judge-correct triplets each stage removes. These two measurements do not cover relations that no paper in the corpus states explicitly, nor triplets whose entities fall outside the CMIP metadata families. Finally, LLM validators may still over-accept plausible but unstated relations or reject relations requiring dispersed evidence or domain knowledge.

\section*{Ethical Considerations}

The work uses publicly available publications, bibliographic metadata, and community resources such as ESGF/GCMD, CMIP6 controlled vocabularies, and GeoNames. Human involvement is limited to low-risk validation of evidence-supported triplets from non-sensitive scientific text. We mitigate risks from third-party LLM extraction and validation by requiring explicit textual evidence, checking evidence grounding, and releasing derived resources with appropriate attribution and source-license constraints. We declare no conflicts of interest.

\bibliography{anthology-1,anthology-2,references, custom}
\clearpage
\appendix

\section*{Appendix}
\label{sec:appendix}
\section{Detailed Extraction Methodology}
\label{sec:appendix_methodology}

\begin{figure*}
    \centering
\includegraphics[width=0.95\linewidth]{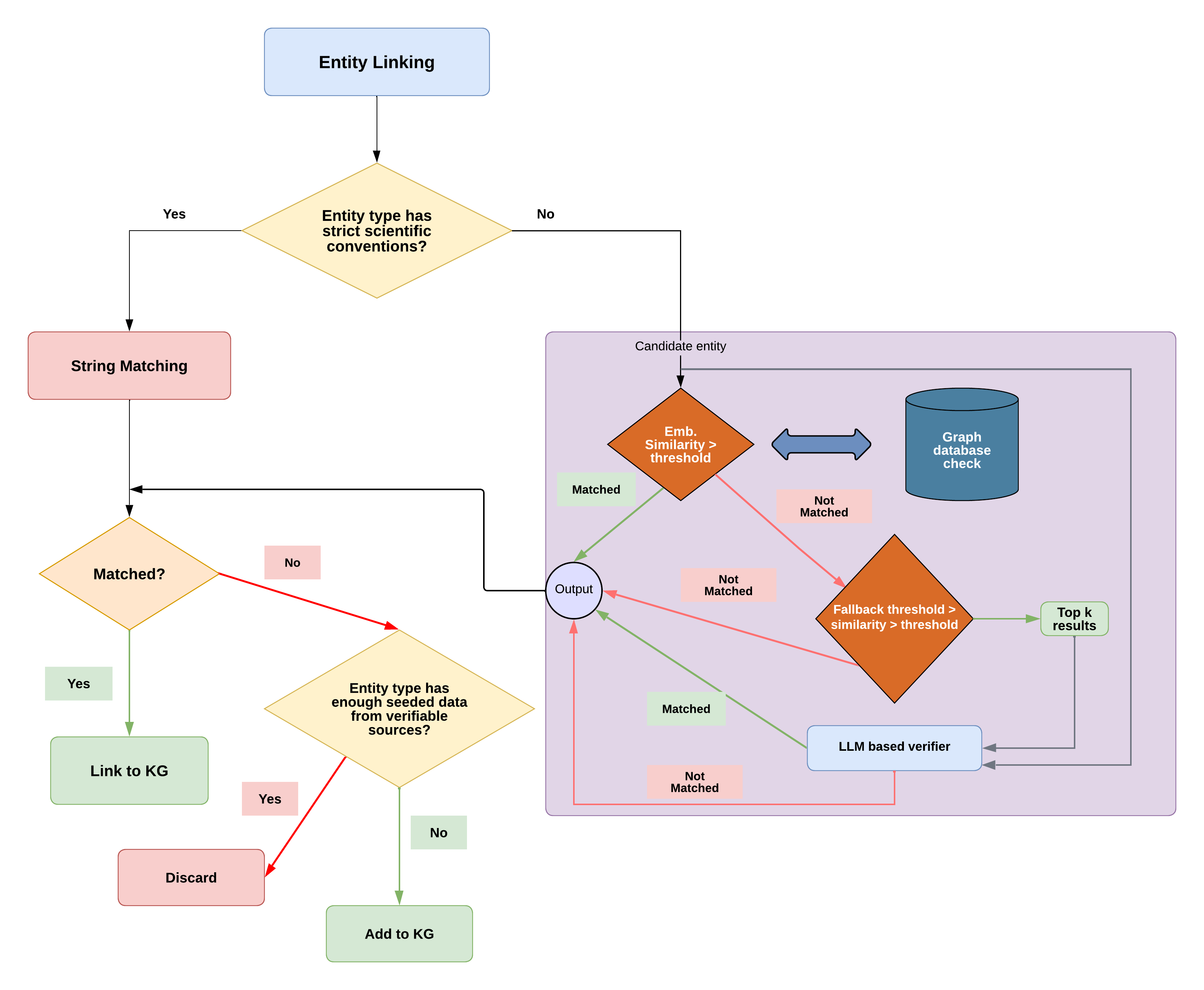}
    \caption{Decision flow for entity linking. Entities are linked via exact string matching when strict naming conventions apply; otherwise, embedding-based similarity with adaptive thresholds and LLM verification is used to link, add, or discard entities from the KG.}
    \label{fig:entity_linking}
\end{figure*}

This appendix provides the specific prompt structures and decision logic used in the agents described in Figure~\ref{fig:info_extraction} and Figure~\ref{fig:entity_linking}. The detailed extraction pipeline is provided in Figure~\ref{fig:detailed_system_pipeline}.

\subsection{Paper Screening Protocols}
\label{subsec:app_screening}

\begin{figure*}
    \centering
    \includegraphics[width=0.95\linewidth]{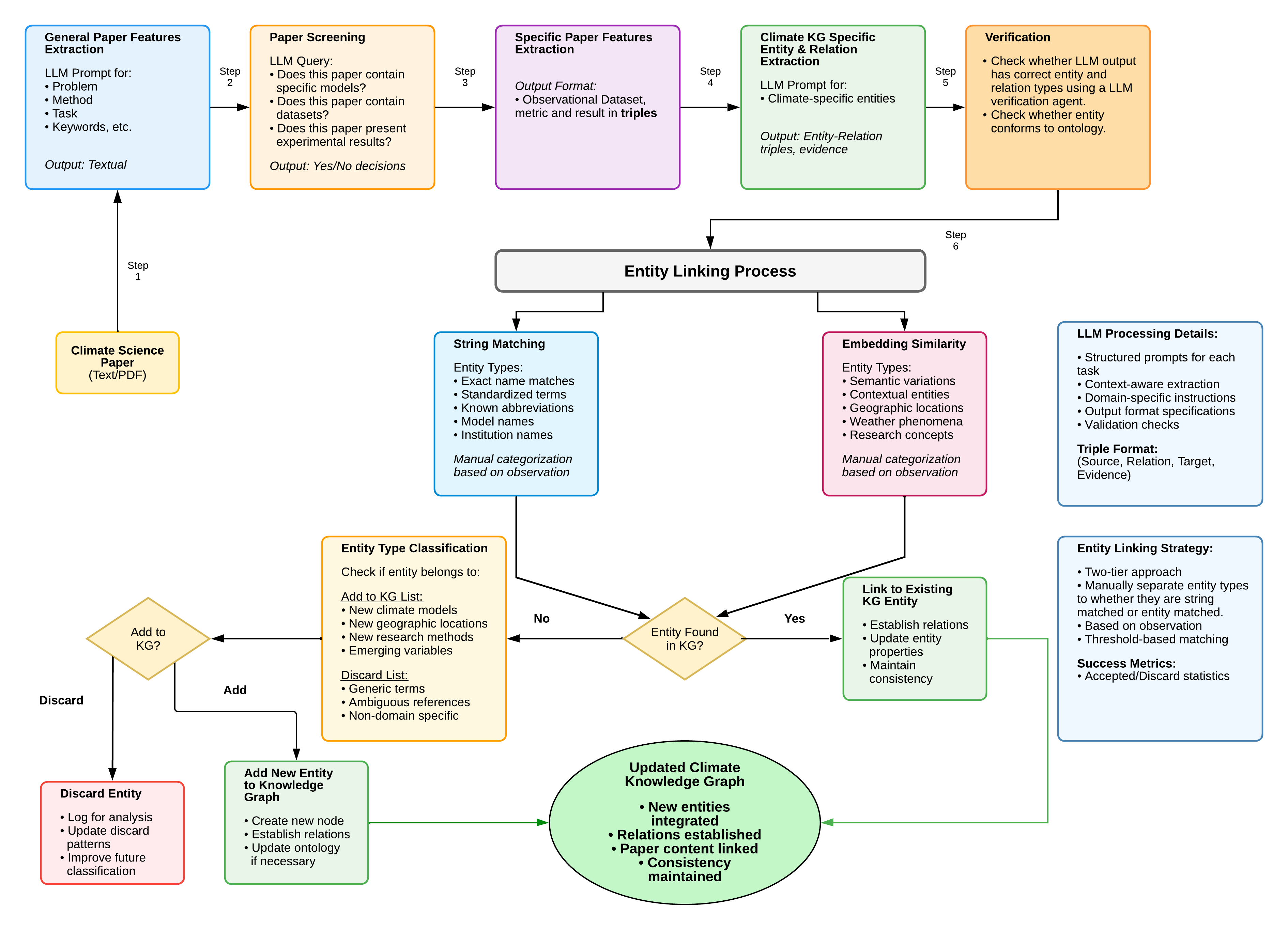}
    \caption{Overview of the ontology-guided, evidence-based KG construction pipeline. The workflow proceeds in three phases: (1) \textbf{Paper Screening and Feature Extraction}, where documents are filtered for empirical relevance (Step 1--3); (2) \textbf{Extraction and Verification}, where an LLM extracts entity-relation triples supported by explicit textual evidence and verifies ontology conformance (Step 4--5); and (3) \textbf{Hybrid Entity Linking}, which routes entities to either deterministic string matching or semantic embedding similarity (Step 6) to determine whether to discard the entity or integrate it into the Updated Climate KG.}
    \label{fig:detailed_system_pipeline}
\end{figure*}

To ensure the KG is populated only with empirical climate research rather than review articles or opinion pieces, the paper screening agent (Step 2 in Figure~\ref{fig:detailed_system_pipeline}) utilizes a strict boolean filtering criteria. The LLM is prompted to answer three specific questions regarding the input document. A paper is retained only if it yields a positive classification for at least one of the following criteria:

\begin{itemize}
    \item \textbf{Model Presence:} ``Does this paper contain climate models (e.g., CMIP6, ESGF-listed models)?''
    \item \textbf{Dataset Availability:} ``Does this paper contain or reference specific observational or simulation datasets?''
    \item \textbf{Empirical Nature:} ``Does this paper present quantitative experimental results?''
\end{itemize}

The actual prompt is provided in the Section \ref{subsubsec:paper_screening_prompt}.

\begin{figure}[h]
    \centering
    \includegraphics[width=0.95\linewidth]{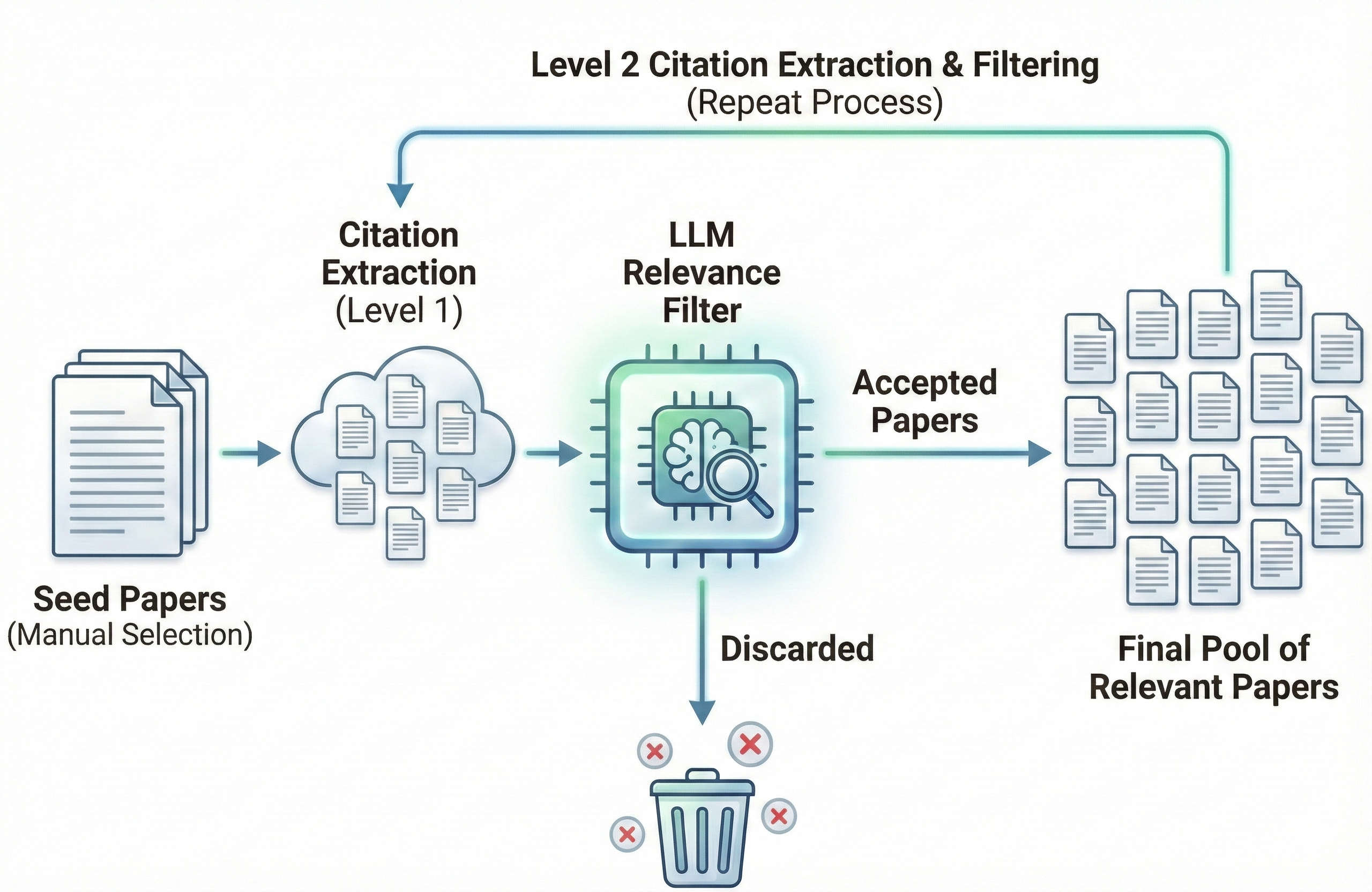}
    \caption{Pipeline for selection of relevant scientific publications. The pipeline starts with a manual seed of highly relevant scientific publications, followed by iterative citation mining and LLM-based filtering to build a high-quality corpus.}
    \label{fig:paper_selection}
\end{figure}

\subsection{Document level features}
\label{subsec:document_level_features}
Given a scientific paper as a PDF document, we first obtain document-level features, such as \textit{title}, \textit{authors}, \textit{publication venue}, and \textit{abstract}, from metadata provided by OpenAlex. Similar to \cite{lan_nlp-akg_2025}, we first identify general features like \textit{problem}, \textit{method}, \textit{task}, which are common to every scientific publication and domain-specific features guided by the target ontology, including entities and concepts relevant to the domain (e.g., models, datasets, variables, performance).

\subsection{Graph Initialization Details}
\label{subsec:app_graph_initialization}

We initialize the KG from structured ESGF dataset metadata (see Section~\ref{subsec:graph_initialization} for the exact snapshot and endpoint), where each record describes a climate simulation using attributes such as experiment identifiers, source model IDs, variables, institutions, and spatial coverage. Dataset-level attributes are deterministically converted into typed ontology entities, while semantic relationships such as \texttt{usedInExperiment}, \texttt{producesVariable}, and \texttt{coversRegion} are instantiated as graph edges. Geospatial bounding boxes are normalized and linked to GeoNames entities, and CMIP6 controlled vocabulary terms are used to enrich model genealogy and component information. The concepts like Ocean circulation, Teleconnection, Instrument and Platform are initialized from GCMD. This initialization grounds later LLM extractions in standardized identifiers; for example, an extracted mention of \emph{precipitation} can be linked to the canonical ESGF variable \emph{pr} rather than requiring the LLM to generate the identifier directly.

\subsection{Entity Linking Categorization Strategy}
\label{subsec:app_linking_strategy}

As illustrated in the Entity Linking Process (Step 6), we employ a hybrid linking strategy that routes entities to either deterministic string matching or semantic embedding similarity based on the entity type. Table~\ref{tab:linking_logic} details the categorization logic used to select the linking method.

We use string matching for rigid identifiers such as climate model names because these entities follow standardized naming conventions where small surface differences can indicate different entities. For example, \texttt{ACCESS-ESM1} and \texttt{ACCESS-ESM1.5} may have high embedding similarity but refer to distinct models, making exact or rule-based matching safer than embedding distance.

\begin{table}[h]
\centering
\small
\caption{Categorization strategy for routing extracted entities to the appropriate linking mechanism.}
\label{tab:linking_logic}
\begin{tabularx}{\linewidth}{@{}>{\raggedright\arraybackslash}X>{\raggedright\arraybackslash}X@{}}
\toprule
\textbf{String Matching Path} & \textbf{Embedding Similarity Path} \\
\textit{Strict rigid identifiers} & \textit{Semantic or contextual concepts} \\
\midrule
Exact name matches; standardized taxonomy terms; known domain abbreviations; official model names (e.g., CESM2); institution names. &
Semantic variations of variables; context-dependent entities; geographic locations; weather phenomena descriptions; abstract research concepts. \\
\bottomrule
\end{tabularx}
\end{table}

\subsection{Entity Linking Threshold Sensitivity Analysis}
\label{subsec:app_entity_linking_threshold_sensitivity}

Table~\ref{tab:entity_linking_threshold_sensitivity} reports the sensitivity analysis used to select the embedding thresholds for entity linking. Scores are top-1 embedding cosine similarities, and LLM decisions are used as the match/reject signal for the sampled subset. The high match rate in $[0.80, 0.85)$ supports auto-linking at $\tau = 0.80$, while nontrivial match rates in $[0.55, 0.80)$ justify using $\tau_f = 0.55$ as the lower bound for LLM fallback instead of discarding those candidates.

\begin{table}[h]
\centering
\small
\setlength{\tabcolsep}{3pt}
\caption{Entity-linking sensitivity analysis by top-1 embedding similarity score bin.}
\label{tab:entity_linking_threshold_sensitivity}
\begin{tabular}{lrrrr}
\toprule
\textbf{Score Bin} & \textbf{Matched} & \textbf{Rejected} & \textbf{Total} & \textbf{Rate} \\
\midrule
$[0.50,\;0.55)$ & 186 & 185 & 371 & 50.1\% \\
$[0.55,\;0.60)$ & 281 & 205 & 486 & 57.8\% \\
$[0.60,\;0.65)$ & 406 & 283 & 689 & 58.9\% \\
$[0.65,\;0.70)$ & 287 & 232 & 519 & 55.3\% \\
$[0.70,\;0.75)$ & 261 & 170 & 431 & 60.6\% \\
$[0.75,\;0.80)$ & 370 & 144 & 514 & 72.0\% \\
$[0.80,\;0.85)$ & 593 & 43 & 636 & 93.2\% \\
\bottomrule
\end{tabular}
\end{table}

\subsection{Qualifiers and context}
\label{subsec:qualifiers_and_context}
Many climate-science statements hold only under a qualifier: an experiment, an ensemble member, a
forcing, a region, or a period. We handle these structurally rather than by attaching free-text
conditions to a relation. The ontology promotes the qualifier-bearing concepts to first-class
entities, so \texttt{Experiment}, \texttt{SubExperiment}, \texttt{Ensemble}, \texttt{Member},
\texttt{Forcing}, \texttt{Metric}, and \texttt{Result} are nodes in their own right, and a
conditioned statement is expressed as a path through them instead of as a single qualified edge; the
KG contains 35,523 \texttt{RESULT\_HAS\_METRIC} edges of this form. Evidence spans are stored at
paragraph granularity with anaphora resolved, so a qualifier stated next to the claim stays visible
to any reader auditing the triple. A statement whose qualifier the ontology cannot express, on the
other hand, has no licensed type signature and is removed by the conformance agent rather than
entering the KG unconditioned. This is a deliberate trade: it keeps the graph interpretable, and it
is one reason the conformance stage removes textually correct triples.

\subsection{Repeated and cited facts}
\label{subsec:repeated_facts}
The same fact is often asserted by several papers, and papers also restate findings that earlier,
cited work introduced. Because entity linking resolves mentions to canonical KG nodes, such
assertions converge on a single edge rather than on one edge per paper. Each edge accumulates the
identifiers of every paper that asserted it in an \texttt{added\_from\_paper\_id} list, and keeps
each paper's evidence span, so the graph records how many independent documents support a relation
and which passage each of them offers. In the 2,000-publication KG, 11.0\% of the distinct
literature-derived edges carry two or more paper identifiers. What the current design does not do is
distinguish the paper that introduced a finding from the papers that restate it: a citation-aware
provenance model is left for future work.

\subsection{What triples the pipeline filters remove}
\label{subsec:filters_remove}
Relative to raw extractor output, the pipeline discards 14,503 triplets that the judge considers correct. Their distribution across stages shows that this is mostly scope filtering rather than a precision--recall trade: 10,326 (71.2\%) are removed by ontology conformance and are correct statements whose type signature lies outside the schema, so no ontology-grounded KG could load them; 835 (5.8\%) fail evidence grounding, meaning their evidence span cannot be located in the parsed source; and 3,342 (23.0\%) are removed by the validator agent, which as shown above rejects unconfirmed rather than only incorrect assertions. Restricting attention to triplets the schema can express, the pipeline retains 23,085 of 27,262 judge-correct triplets, a within-schema retention of 84.7\%, while raising $\TripletCorrectness$ from 94.05\% to 98.96\%. Measured on the human-annotated subset, within-schema retention is 81.9\%.

\subsection{Downstream Utility}

The constructed KG supports multi-hop scientific queries that combine literature-derived assertions with structured climate metadata. Example questions include:
\begin{itemize}
    \item Which variable is both produced by the REMO2009 regional climate model and measured by Vaisala sondes RS80/RS92/VS41?
    \item Which source component does CanESM5 have that applies to the ocean realm?
    \item What project did the institute that produced the CNRM-ESM2-1 source model also participate in?
    \item What observational dataset was used by the method that relies on ``Last Glacial Maximum to control temperature change'' as a metric?
    \item What is a result of the method that experiments on Antarctic ice core water stable isotopes?
\end{itemize}
These queries illustrate how the KG can connect papers, models, variables, instruments, methods, metrics, results, institutions, and controlled climate vocabularies through auditable paths rather than isolated keyword matches.

\subsection{Other Implementation Details}
\label{subsec:implementation_details}
We use Gemini-2.5-pro for triplet/evidence extraction and LLM-based entity-linking fallback, Gemini-2.5-flash for large-scale validation and evaluation, and Llama 3.3-70B \cite{grattafiori2024llama} for candidate paper selection.

\subsection{Per-Stage Cost and Runtime}
\label{subsec:app_extraction_cost_runtime}

Table~\ref{tab:stage_cost} breaks the cost of the full 2,000-paper run down by stage and places it next to the ablation effect of that stage, so that the price and the benefit of each component can be read together. Costs are estimates at the API prices in force during our runs, and they vary with PDF length and output volume.

Extraction dominates: Gemini-2.5-pro is called roughly five times per paper, which we cost conservatively at about USD 200 per 1,000 papers, and processing 1,000 papers takes about 2--3 hours. This cost is not specific to our design, since any LLM-based KG construction system must read the documents.

The verification stack that follows is close to free. Ontology conformance and evidence validation use no LLM at all: the first is a set-membership test over the licensed (source type, relation, target type) tuples with rule-based normalization, and the second is a fuzzy string match of the evidence span against the parsed document, so both run in CPU minutes at no API cost. The validator agent is the only paid verifier, at one short Gemini-2.5-flash call per surviving triplet: 27,900 calls, approximately 9M input and 0.2M output tokens, or roughly USD 3. Paper screening runs on a locally hosted Llama-3.3-70B and costs GPU time rather than API credit, as does PDF parsing with olmOCR. Entity linking runs local NV-Embed-v2 embeddings and calls an LLM only for candidates that fall in the fallback band $[\tau_f, \tau)$.

Cost across the ablation variants of Table~\ref{tab:stage_ablation} differs only in how many triplets reach the validator agent. Running it without the two deterministic filters raises its call count from 27,900 to 39,964, or roughly USD 4.7, while the extraction cost is common to every variant. Two of the three stages are deterministic and use no LLM, and the third issues one short Gemini-2.5-flash call per surviving triplet, so verification adds roughly 1\% to the cost of extraction, which any LLM-based KG construction system has to pay. That 1\% buys 4.91 points of $\TripletCorrectness$ together with the schema and traceability guarantees.

\begin{table*}[!t]
\centering
\small
\caption{Per-stage cost, runtime, and ablation effect for the full 2,000-publication run. Ablation effect reports the result of leaving that stage out of the full pipeline, which itself yields 23,327 triplets at 98.96\% $\TripletCorrectness$.}
\label{tab:stage_cost}
\begin{tabularx}{\textwidth}{@{}>{\raggedright\arraybackslash}p{0.2\textwidth}>{\raggedright\arraybackslash}X>{\raggedright\arraybackslash}X>{\raggedright\arraybackslash}X@{}}
\toprule
\textbf{Stage} & \textbf{Mechanism} & \textbf{Cost (2,000 papers)} & \textbf{Ablation effect} \\
\midrule
Paper screening & Llama-3.3-70B, local & no API cost, GPU time & corpus selection, not evaluated \\
PDF parsing & olmOCR, local & no API cost, GPU time & prerequisite for evidence grounding \\
Triplet + evidence extraction & Gemini-2.5-pro, $\approx$5 calls/paper & $\approx$USD 400, 2--3 h per 1,000 papers & produces 39,964 triplets at 94.05\% \\
Ontology conformance (O) & deterministic set membership & no API cost, CPU seconds & w/o: 31,095 at 98.88\%, 7,768 untypeable kept \\
Evidence validation (E) & deterministic fuzzy match & no API cost, CPU minutes & w/o: 24,172 at 98.96\%, 845 ungrounded kept \\
Validator agent (V) & Gemini-2.5-flash, 1 call/triplet & $\approx$USD 3 (27,900 calls) & w/o: 27,900 at 94.72\%, $-4.24$ points \\
Entity linking & NV-Embed-v2 local + LLM fallback & GPU time, few fallback calls & links accepted triplets to the KG \\
\bottomrule
\end{tabularx}
\end{table*}

\subsection{Entity Classification and Discard Logic}
\label{subsec:app_discard}

Entities that fail to link to the existing KG via either string matching or embedding similarity (below threshold $\tau_f$) undergo a secondary classification step. This ensures that the graph is enriched with novel but relevant concepts while filtering out noise. The prompt to the LLM is provided in \ref{subsubsec:embedding_fallback_prompt}.

As shown in the pipeline workflow, discarded entities are logged to analyze and update exclusion patterns, creating a feedback loop that refines future classification precision.

\subsection{Climate ontology}

The entity and relation definitions of the climate ontology are given below.
\subsection{Entity Type Definitions}
\begin{itemize}
    \item \textbf{Activity}: A coordinated modeling effort or scientific campaign.
    \item \textbf{ExperimentFamily}: A set of climate model experiments conducted under a common scientific objective (e.g., CMIP experiment families).
    \item \textbf{Experiment}: A specific climate model simulation scenario (e.g., historical, ssp585).
    \item \textbf{SubExperiment}: A variant of a climate model experiment with specific configurations (e.g., initialization, ensemble member, or resolution).
    \item \textbf{Source}: A climate model or system used to generate data (e.g., GFDL-ESM4).
    \item \textbf{SourceType}: The classification of the model.
    \item \textbf{SourceComponent}: A sub-model (e.g., atmosphere, ocean, land) within a coupled climate model; may also function as a standalone model (e.g., ACCESS-OM2).
    \item \textbf{PhysicalScheme}: A representation of a physical process within a climate model (e.g., cloud scheme).
    \item \textbf{PhysicalFeature}: A structural characteristic of a climate model’s design (e.g., terrain-following grid).
    \item \textbf{SimulationType}: Type of simulation performed (e.g., transient, equilibrium).
    \item \textbf{Metric}: Quantitative measure of model performance (e.g., climate sensitivity).
    \item \textbf{Project}: A broader initiative under which models or experiments are conducted (e.g., CMIP6).
    \item \textbf{Institute}: Organization responsible for developing models or running simulations.
    \item \textbf{Variable}: Scientific quantities output by models (e.g., temperature, precipitation).
    \item \textbf{Realm}: Component of the Earth system (e.g., atmosphere, ocean).
    \item \textbf{Frequency}: Temporal resolution of model output (e.g., daily, monthly).
    \item \textbf{Resolution}: Spatial resolution or grid size of the data.
    \item \textbf{Ensemble}: A collection of model runs differing in initial conditions or configurations.
    \item \textbf{Member}: An individual member of an ensemble.
    \item \textbf{MIPEra}: A major generation or version of coordinated experiments (e.g., CMIP5, CMIP6).
    \item \textbf{RCM}: A regional climate model used for fine-resolution regional simulations.
    \item \textbf{Domain}: Geographical coverage of a model.
    \item \textbf{Continent}: A large continuous landmass (e.g., Asia, Africa).
    \item \textbf{Country}: A sovereign state or territory (e.g., India, USA).
    \item \textbf{Country\_Subdivision}: Administrative units within countries (e.g., California).
    \item \textbf{City}: Urban locality (e.g., Paris).
    \item \textbf{No\_Country\_Region}: Areas not under country jurisdiction (e.g., open ocean).
    \item \textbf{Instrument}: Device used to observe environmental variables (e.g., radiometer).
    \item \textbf{Platform}: Physical carrier for an instrument (e.g., satellite).
    \item \textbf{Weather\_Event}: Specific events such as storms or droughts.
    \item \textbf{Teleconnection}: Large-scale climate patterns (e.g., ENSO, NAO).
    \item \textbf{Ocean\_Circulation}: Movements of ocean waters (e.g., AMOC).
    \item \textbf{Natural\_Hazard}: Geophysical events impacting systems (e.g., tsunami, earthquake).
    \item \textbf{NaturalHazardType}: The specific type of natural hazard (e.g., Flood, Tornado, Wildfire).
    \item \textbf{ClimateZone}: Classification of climate based on temperature and precipitation patterns (Tropical, Dry, Temperate, Continental, Polar).
\end{itemize}

\subsection{Relation Definitions}
\begin{itemize}
    \item \textbf{INCLUDES\_EXPERIMENT}: Links an ExperimentFamily to its constituent Experiments.
    \item \textbf{HAS\_SUB\_EXPERIMENT}: Connects an Experiment to a refining SubExperiment.
    \item \textbf{IS\_OF\_TYPE}: Connects a Source or Natural\_Hazard to its corresponding type.
    \item \textbf{APPLIES\_TO\_REALM}: Associates a Source, SourceComponent, or Experiment with a Realm.
    \item \textbf{FOCUSES\_ON\_REALM}: Indicates the Realm emphasized by an Activity.
    \item \textbf{PARTICIPATED\_IN}: Connects an Institute to a Project.
    \item \textbf{GENERATED\_BY\_ACTIVITY}: Indicates that a Source was produced as part of an Activity.
    \item \textbf{PRODUCED\_BY\_INSTITUTE}: Links a Source to the responsible Institute.
    \item \textbf{BELONGS\_TO\_MIP\_ERA}: Indicates association with a specific MIP Era.
    \item \textbf{USED\_IN\_EXPERIMENT}: Links a Source or RCM to an Experiment.
    \item \textbf{PRODUCES\_VARIABLE}: Specifies Variables output by a Source or SourceComponent.
    \item \textbf{HAS\_SUBSEQUENT\_VERSION}: Links a Source or SourceComponent to a newer version.
    \item \textbf{INHERITED\_FROM}: Indicates derivation or inheritance between Sources.
    \item \textbf{HAS\_SOURCE\_COMPONENT}: Connects a Source to its SourceComponents.
    \item \textbf{USES\_PHYSICAL\_SCHEME}: Indicates a PhysicalScheme employed by a Source.
    \item \textbf{HAS\_PHYSICAL\_FEATURE}: Links a Source to a defining PhysicalFeature.
    \item \textbf{HAS\_SIMULATION\_TYPE}: Connects a Source to a SimulationType.
    \item \textbf{HAS\_METRIC}: Associates a Source with an evaluation Metric.
    \item \textbf{PART\_OF\_PROJECT}: Indicates membership in a Project.
    \item \textbf{ASSOCIATED\_WITH\_MEMBER}: Connects a Source to an Ensemble Member.
    \item \textbf{SAMPLED\_AT\_FREQUENCY}: Indicates output Frequency of a Source.
    \item \textbf{HAS\_SPATIAL\_RESOLUTION}: Specifies the spatial Resolution of an entity.
    \item \textbf{PART\_OF\_ENSEMBLE}: Links a Source or SubExperiment to an Ensemble.
    \item \textbf{USES\_FORCING}: Indicates forcings used in an Experiment.
    \item \textbf{INCLUDES\_ENSEMBLE\_MEMBER}: Connects an Experiment to its Ensemble members.
    \item \textbf{DRIVEN\_BY\_SOURCE}: Indicates that an RCM is driven by a Source.
    \item \textbf{COVERS\_DOMAIN}: Specifies the Domain covered by an RCM or Project.
    \item \textbf{MEASURES\_VARIABLE}: Links an Instrument to the Variable it measures.
    \item \textbf{MOUNTED\_ON}: Indicates the Platform carrying an Instrument.
    \item \textbf{PERFORMED\_BY\_INSTITUTE}: Indicates which Institute performed an Experiment.
    \item \textbf{COVERS\_REGION}: Specifies the region spanned by an RCM.
    \item \textbf{IMPACTED\_REGION}: Indicates the region affected by a Weather\_Event.
    \item \textbf{INFLUENCES\_REGION}: Links a Teleconnection to influenced regions.
    \item \textbf{OBSERVED\_IN\_REGION}: Indicates regions where an Ocean\_Circulation is observed.
    \item \textbf{AFFECTED\_REGION}: Indicates regions affected by a Natural\_Hazard.
    \item \textbf{IN\_CONTINENT}: Connects a Country to its Continent.
    \item \textbf{IN\_COUNTRY}: Connects a Country\_Subdivision or City to its Country.
    \item \textbf{HAS\_CLIMATE\_ZONE}: Links a No\_Country\_Region to a ClimateZone.
\end{itemize}

\subsection{Scientific publications ontology}
\label{subsec:app_scientific_paper_ontology}
The scientific publications ontology is inspired from \cite{lan_nlp-akg_2025} and modified for climate domain.
\paragraph{Entity Definitions}
\begin{itemize}
    \item \textbf{Paper}: A scholarly publication that reports research contributions, methods, experiments, and results.
    
    \item \textbf{Field}: A broad research area or discipline to which a paper contributes (e.g., information retrieval, computer vision).
    
    \item \textbf{Problem}: A specific research challenge or question that the paper aims to address.
    
    \item \textbf{Method}: A technical approach, algorithm, or procedure proposed or applied to solve a problem.
    
    \item \textbf{Innovation}: A novel idea, contribution, or improvement introduced by the paper relative to prior work.
    
    \item \textbf{Task}: A well-defined computational or scientific objective used to evaluate methods (e.g., classification, forecasting).
    
    \item \textbf{Venue}: The conference, journal, or workshop where the paper is published.
    
    \item \textbf{Author}: An individual who contributed intellectually to the creation of the paper.
    
    \item \textbf{Keyword}: A salient term or phrase summarizing key topics or concepts of the paper.
    
    \item \textbf{ObservationalDataset}: A dataset used for experimentation or evaluation, typically derived from real-world or empirical observations.
    
    \item \textbf{Metric}: A quantitative measure used to evaluate the performance or quality of a method or model.
    
    \item \textbf{Result}: An experimental outcome or reported finding, often associated with a metric value.
    
\end{itemize}

\paragraph{Relation Types}
The scientific-publication ontology contains paper-level relations for venue, authorship, keywords, model use, metrics, results, and paper mentions; method- and task-level relations for problems, innovations, tasks, metrics, observational datasets, and results; model- and result-level relations for observational datasets, metrics, and reported results; and LLM-extracted ontology-aligned relations stored with provenance (\texttt{added\_from\_paper\_id}).

\subsection{Final KG statistic}
After KG enrichment the final statistic is provided in Table~\ref{tab:entity_type_counts} and Table~\ref{tab:relation_type_counts}.

\subsection{Justification for our selection of climate domain ontology}
\label{subsec:app_justification_ontology}

In order to benefit from established scientific terminology and ensure interoperability with existing climate data infrastructures, our domain ontology is constructed by unifying multiple authoritative sources of climate terminology. Specifically, we integrate standardized climate model names, institutions, experiments, and variables from ESGF metadata\footnote[1]{\url{https://esgf.github.io/}} and the GCMD taxonomy defined in \cite{pan_taxonomy-driven_2025}, which together provide vetted, community-accepted identifiers and a widely adopted semantic taxonomy for Earth system science. To capture internal model structure and configuration details that are not explicitly represented in ESGF metadata, we further incorporate climate model component information from the CMIP6 Controlled Vocabulary repository\footnote[2]{\url{https://github.com/WCRP-CMIP/CMIP6_CVs}}, enabling canonical representation of model components and experiments. Finally, to provide unambiguous spatial grounding, geographic entities such as regions and countries are linked to GeoNames identifiers, supporting spatially contextualized relations (e.g., model evaluation over specific regions) and facilitating integration with external KGs that rely on standardized geographic identifiers.

\subsection{Ontology Conformance Agent Specification}
\label{subsec:app_conformance_spec}

The ontology conformance agent is fully deterministic: it uses no LLM, no external reasoner, and no
learned component. The curated ontology is compiled into a set $\mathcal{L}$ of 35 licensed tuples
(source type, target type, relation), and a triple conforms if and only if it passes both of the
following tests.

\paragraph{1. Licensing.} The triple's type signature must be a member of $\mathcal{L}$:
\[
(c_h, c_t, r) \in \mathcal{L}.
\]
The test is a constant-time set lookup after case and whitespace normalization of the type names.
For example, \texttt{(Source, SourceComponent, HAS\_SOURCE\_COMPONENT)} is licensed, whereas the
reversed extraction \texttt{(SourceComponent, Source, HAS\_SOURCE\_COMPONENT)} is not, which is the
case discussed in Section~\ref{subsec:ie_pipeline}. Licensing is what makes relation direction and
argument order enforceable without asking a model to check them.

\paragraph{2. Semantic boundary guards.} Licensing constrains types, not the values that fill them,
so a second set of rules rejects values that carry the right type name but the wrong kind of
content. The guards are built from CMIP6 controlled-vocabulary lists and regular expressions, and
they fire on the three types where the extractor confuses categories most often:

\begin{itemize}
\item \texttt{SourceComponent} must not be a realm label. A value in the CMIP6 realm vocabulary
(\emph{atmos}, \emph{ocean}, \emph{seaice}, \emph{landice}, and their spelled-out variants) is
rejected, because the licensed way to state that fact is
\texttt{(SourceComponent, Realm, APPLIES\_TO\_REALM)}.
\item \texttt{PhysicalScheme} must show a scheme cue. Either the value or its evidence paragraph has
to contain a cue word (\emph{scheme}, \emph{parameterization}, \emph{closure}, \emph{formulation},
\emph{algorithm}, \emph{physics package}) or a known scheme name (\emph{Tiedtke},
\emph{Zhang-McFarlane}, \emph{Arakawa-Schubert}, \emph{Betts-Miller}, \emph{KPP}, and others).
\item \texttt{PhysicalFeature} must not be a value of another category. Ensemble member identifiers
matching \texttt{r\{d\}i\{d\}p\{d\}f\{d\}}, resolutions such as ``100 km'' or ``1.5 degrees'',
experiment or scenario labels such as \emph{historical} and \texttt{ssp585}, activity and project
labels such as \emph{CMIP6} and \emph{CORDEX}, and common variable names such as \emph{tas} and
\emph{pr} are all rejected.
\end{itemize}

Because both tests are exact and side-effect free, the stage is reproducible, auditable, and free of
API cost, and its verdict is stored per triple so that the decision can be revisited without
re-running any model. Widening the ontology therefore amounts to adding tuples to $\mathcal{L}$,
which is the human-in-the-loop path we suggest for the correct-but-untypeable triples reported in
Section~\ref{subsec:recall}.

\subsection{Per-Stage Examples}
\label{subsec:app_stage_examples}

The following examples are taken verbatim from the 2,000-publication run and illustrate what each stage keeps and what it removes. Evidence spans are abbreviated.

\paragraph{Kept by every stage.}
(\emph{severe floods}, \texttt{Natural\_Hazard}) \texttt{AFFECTED\_REGION} (\emph{central and western Africa}, \texttt{No\_Country\_Region}), with evidence ``In 2012, central and western Africa was hit by severe floods which affected 1,538,242 people and caused 340 deaths as of September of that year.'' The type signature is licensed by the ontology, the span occurs in the parsed document, and the validator agent confirms that the span states the relation.

\paragraph{Removed by ontology conformance, yet textually correct.}
(\emph{MRI-CGCM3}, \texttt{Source}) \texttt{PRODUCES\_VARIABLE} (\emph{u-component of wind}, \texttt{Variable}). The paper does state this, and the judge marks it correct, but the ontology licenses \texttt{PRODUCES\_VARIABLE} only from \texttt{SourceComponent} to \texttt{Variable}, because in the ESGF data model it is a model component that produces a variable. The triplet has no place in the schema and is therefore not loadable, which is the case for 93.4\% of what this stage removes.

\paragraph{Removed by evidence validation.}
(\emph{Integrated Forecasting System (IFS)}, \texttt{Source}) \texttt{HAS\_PHYSICAL\_FEATURE} (\emph{137 model levels}, \texttt{PhysicalFeature}). The claim is plausible and the wording resembles the paper, but the span the extractor returned cannot be matched in the parsed document above the similarity threshold, so the triplet loses its audit trail and is dropped. Such spans are frequently reconstructed from tables and figure captions, which the parser renders differently.

\paragraph{Removed by the validator agent, yet accepted by human annotators.}
(\emph{FIO-ESM v2.0}, \texttt{Source}) \texttt{HAS\_SOURCE\_COMPONENT} (\emph{CLM4.0}, \texttt{SourceComponent}), with evidence ``The land surface model is upgraded from the Community Land Model version 3.5 (CLM3.5; Oleson et al., 2008) to CLM4.0 (Lawrence et al., 2011) with the same horizontal resolution as CAM5.'' The span states that the land model is upgraded to CLM4.0 without naming the parent model, so the validator cannot confirm the attachment from the span alone, while annotators who read the surrounding paper accept it. This is the conservative behaviour quantified in Section~\ref{subsec:recall}: 79\% of the validator's rejections in the annotated sample are judged valid by the human majority.

\subsection{Evidence Verification Against Parsed PDF and Content and Hallucination Analysis}

We assess the factual reliability of LLM-extracted evidence by verifying it against the original PDF content parsed using the Olmocr v2 \cite{poznanski_olmocr_2025} parser. For each extracted triple, the associated evidence span generated by the LLM is aligned with the parsed PDF text. An evidence instance is considered \emph{correct} if the cited text can be unambiguously located in the parsed content and faithfully supports the extracted relation; otherwise, it is marked as a hallucination.

For each extracted triple, the extractor also outputs a supporting evidence span. We compute the textual similarity between the extracted evidence span and the source document using the RapidFuzz library. RapidFuzz provides normalized string-matching scores in the range $[0,100]$, where higher values indicate stronger lexical overlap. In our experiments, a triple is considered evidence-grounded if its similarity score exceeds a predefined threshold (70\%, 75\%, or 80\%). Triples failing this criterion are flagged as hallucinations.

Table \ref{tab:hallucination_rate_evidence} reports results when all evidence spans are considered (including those derived from tables and figures), hallucination rates range from $7.74\%$ at an $80\%$ threshold to $6.14\%$ at a $70\%$ threshold. As expected, relaxing the similarity threshold admits more borderline matches and thus reduces the measured hallucination rate. 

However, a large portion of the mismatch is due to the way Olmocr V2 and Gemini-2.5-pro interpret tables and figures. When we do not consider extracted texts explaining table and figure portions of the document the hallucination rates drop sharply—down to $1.44\% (80\%)$, $1.04\% (75\%)$, and $0.63\% (70\%)$. This result highlights that the hallucination rate of Gemini-2.5-pro is relatively low to extract accurate text-spans from documents. 

\begin{table}[t]
\centering
\small
\caption{Hallucination analysis of an LLM-based KG extractor agent with evidence.}
\label{tab:hallucination_rate_evidence}
\begin{tabularx}{\linewidth}{@{}>{\raggedright\arraybackslash}Xcc@{}}
\toprule
\textbf{Experimental setting} & \textbf{Threshold} & \textbf{Rate (\%)} \\
\midrule
Evidence span & 80\% & 7.741 \\
Evidence span & 75\% & 6.793 \\
Evidence span & 70\% & 6.137 \\
Evidence span without table/figure interpretations & 80\% & 1.443 \\
Evidence span without table/figure interpretations & 75\% & 1.035 \\
Evidence span without table/figure interpretations & 70\% & 0.626 \\
\bottomrule
\end{tabularx}
\end{table}

\subsection{AI use disclosure}
We used AI assistants, including GPT-5.4, only for limited manuscript-support tasks: improving wording/clarity and helping design or refine the pipeline figure. The scientific ideas, methodology, experiments, analysis, and final claims were developed and verified by the authors. No AI system was used to generate experimental results or make autonomous research decisions.

\subsection{Annotation guidelines and instructions}

\subsubsection{Annotator Qualifications and Selection}
All 3 human experts hold PhD in closely related field. They were recruited through a third-party online professional services platform. The experts compensation was according to current standards. They were recruited for the specific task of annotation and were preinformed about their task and the compensation they would receive. The participants were informed how the annotation was going to be used. The annotations began only after both parties agreed on the terms.

\subsubsection{Annotation Process Overview} 
Annotators are tasked with validating extracted knowledge graph triples with respect to their supporting textual evidence. Given the relatively lightweight nature of this task, annotators are not required to perform open-ended information extraction or domain discovery. Instead, they are provided only with the ontology definitions, including entity types and relation descriptions, to guide consistent judgments.

For each candidate triple, annotators are shown: (i) the structured triple, (ii) the corresponding evidence paragraph extracted from the source document, and (iii) the relevant ontological definitions. Annotators are instructed to verify whether the evidence explicitly supports the triple and whether the triple conforms to the ontology’s domain and range constraints. No additional domain expertise beyond the provided definitions is assumed, and annotators are not asked to infer unstated facts or resolve ambiguities beyond the given text.

\begin{figure*}
    \centering
    \includegraphics[width=0.8\linewidth]{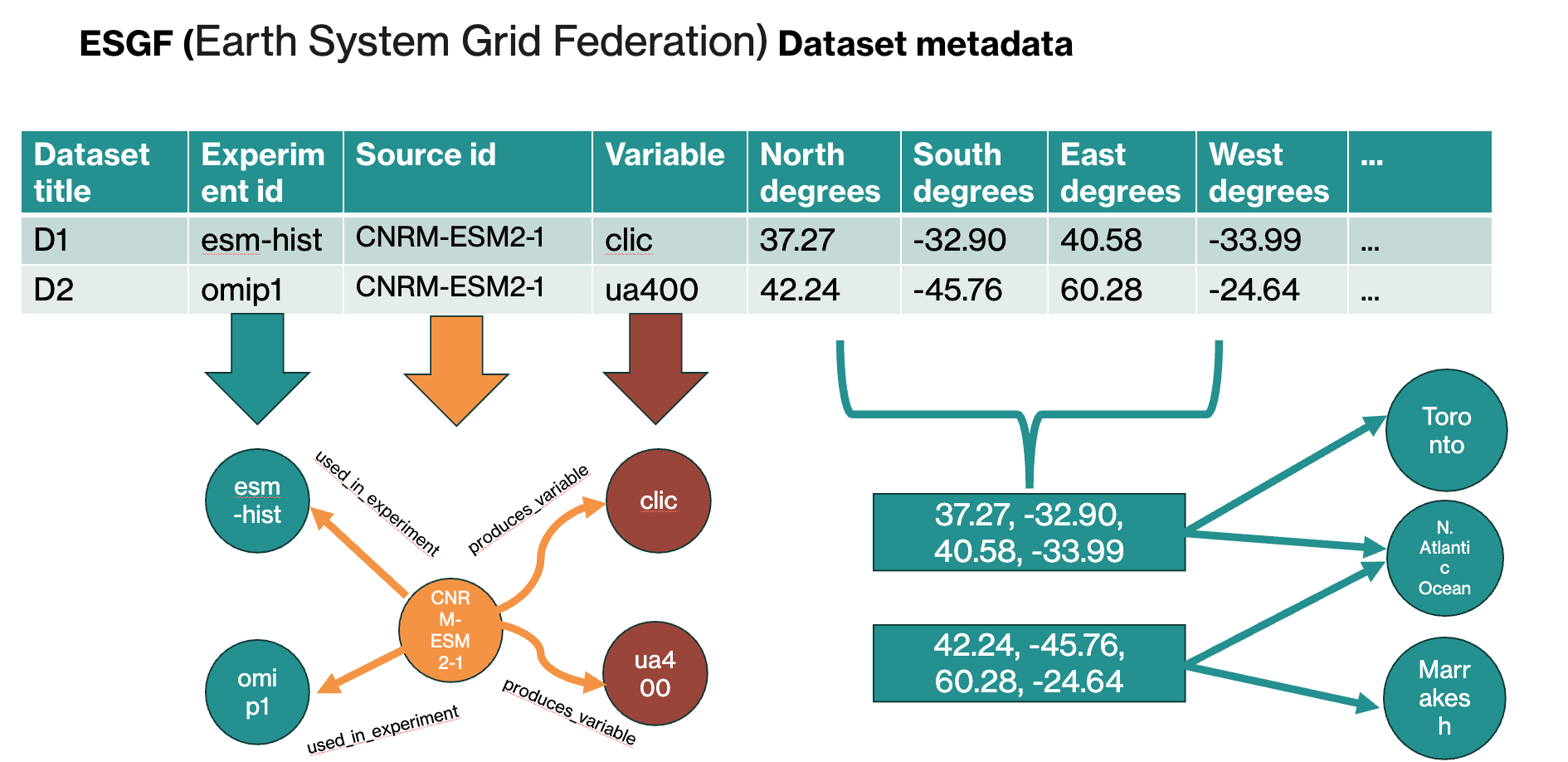}
    \caption{A small subsection of structured ESGF metadata converted to KG based on ontology specifications mapped with geolocation data from Geonames.}
    \label{fig:esgf_metadata_to_kg}
\end{figure*}

\begin{table*}[t]
\centering
\small
\setlength{\tabcolsep}{6pt}
\caption{Distribution of entity types in the constructed KG.}
\label{tab:entity_type_counts}
\begin{tabular*}{0.92\textwidth}{@{\extracolsep{\fill}}lrlr@{}}
\toprule
\textbf{Entity Type} & \textbf{Count} & \textbf{Entity Type} & \textbf{Count} \\
\midrule
City & 30{,}062 & Ensemble & 578 \\
Result & 16{,}207 & Experiment & 481 \\
Author & 10{,}437 & Source & 394 \\
EvaluationMetric & 8{,}430 & Venue & 373 \\
No\_Country\_Region & 6{,}965 & Country & 252 \\
PhysicalScheme & 6{,}568 & Forcing & 204 \\
Metric & 6{,}144 & Institute & 132 \\
Variable & 6{,}056 & SubExperiment & 63 \\
Member & 5{,}929 & NaturalHazardType & 48 \\
Country\_Subdivision & 3{,}893 & RCM & 42 \\
PhysicalFeature & 3{,}236 & Domain & 39 \\
Instrument & 3{,}045 & Project & 37 \\
ObservationalDataset & 2{,}496 & Teleconnection & 27 \\
SourceComponent & 2{,}440 & Activity & 26 \\
Keyword & 2{,}344 & Ocean\_Circulation & 20 \\
Platform & 2{,}326 & GridLabel & 17 \\
Innovation & 2{,}000 & Frequency & 16 \\
Method & 2{,}000 & SourceType & 16 \\
Paper & 2{,}000 & Resolution & 15 \\
Problem & 2{,}000 & Realm & 14 \\
Task & 2{,}000 & ExperimentFamily & 9 \\
Weather\_Event & 1{,}645 & Continent & 7 \\
Field & 1{,}280 & ClimateZone & 5 \\
SimulationType & 840 & MIPEra & 4 \\
Natural\_Hazard & 828 & \textbf{Total} & \textbf{133{,}990} \\
\bottomrule
\end{tabular*}
\end{table*}

\begin{table*}[t]
\centering
\small
\setlength{\tabcolsep}{6pt}
\caption{Distribution of scientific paper relation types in the constructed KG.}
\label{tab:relation_type_counts}
\begin{tabular*}{0.82\textwidth}{@{\extracolsep{\fill}}lr@{}}
\toprule
\textbf{Relation Type} & \textbf{Count} \\
\midrule
PAPER\_MENTIONS & 96{,}100 \\
RESULT\_HAS\_METRIC & 35{,}522 \\
PAPER\_HAS\_RESULT & 29{,}879 \\
METHOD\_HAS\_RESULT & 29{,}879 \\
PAPER\_HAS\_KEYWORD & 22{,}644 \\
PAPER\_AUTHORED\_BY & 14{,}853 \\
PAPER\_USES\_METRIC & 8{,}720 \\
METHOD\_USES\_METRIC & 8{,}720 \\
TASK\_USES\_METRIC & 8{,}720 \\
MODEL\_USES\_METRIC & 4{,}797 \\
MODEL\_HAS\_RESULT & 4{,}362 \\
PAPER\_USES\_MODEL & 2{,}800 \\
PAPER\_EXPERIMENTS\_ON\_OBSERVATIONAL\_DATASET & 2{,}749 \\
METHOD\_EXPERIMENTS\_ON\_OBSERVATIONAL\_DATASET & 2{,}749 \\
TASK\_EXPERIMENTS\_ON\_OBSERVATIONAL\_DATASET & 2{,}749 \\
PAPER\_BELONGS\_TO\_FIELD & 2{,}000 \\
PAPER\_SOLVES\_PROBLEM & 2{,}000 \\
PAPER\_APPLIES\_METHOD & 2{,}000 \\
PAPER\_HAS\_INNOVATION & 2{,}000 \\
PAPER\_WORKS\_ON\_TASK & 2{,}000 \\
PAPER\_PUBLISHED\_IN\_VENUE & 2{,}000 \\
METHOD\_SOLVES\_PROBLEM & 2{,}000 \\
TASK\_FACES\_PROBLEM & 2{,}000 \\
METHOD\_WORKS\_ON\_TASK & 2{,}000 \\
METHOD\_HAS\_INNOVATION & 2{,}000 \\
MODEL\_EXPERIMENTS\_ON\_OBSERVATIONAL\_DATASET & 876 \\
\bottomrule
\end{tabular*}
\end{table*}

\begin{table*}[t]
\centering
\small
\setlength{\tabcolsep}{6pt}
\caption{Distribution of climate relation types in the constructed KG.}
\label{tab:climate_relation_type_counts}
\begin{tabular*}{0.72\textwidth}{@{\extracolsep{\fill}}lr@{}}
\toprule
\textbf{Relation Type} & \textbf{Count} \\
\midrule
COVERS\_REGION & 354{,}837 \\
IN\_COUNTRY & 33{,}923 \\
PRODUCES\_VARIABLE & 31{,}024 \\
ASSOCIATED\_WITH\_MEMBER & 11{,}903 \\
HAS\_SPATIAL\_RESOLUTION & 6{,}708 \\
APPLIES\_TO\_REALM & 4{,}031 \\
INCLUDES\_ENSEMBLE\_MEMBER & 3{,}069 \\
PERFORMED\_BY\_INSTITUTE & 2{,}762 \\
USED\_IN\_EXPERIMENT & 2{,}019 \\
PART\_OF\_PROJECT & 1{,}048 \\
USES\_FORCING & 986 \\
SAMPLED\_AT\_FREQUENCY & 828 \\
HAS\_SOURCE\_COMPONENT & 734 \\
GENERATED\_BY\_ACTIVITY & 623 \\
MEASURES\_VARIABLE & 607 \\
IS\_OF\_TYPE & 542 \\
PRODUCED\_BY\_INSTITUTE & 503 \\
HAS\_METRIC & 465 \\
MOUNTED\_ON & 422 \\
INCLUDES\_EXPERIMENT & 328 \\
BELONGS\_TO\_MIP\_ERA & 318 \\
HAS\_PHYSICAL\_FEATURE & 298 \\
PARTICIPATED\_IN & 213 \\
IN\_CONTINENT & 210 \\
FOCUSES\_ON\_REALM & 170 \\
DRIVEN\_BY\_SOURCE & 139 \\
COVERS\_DOMAIN & 125 \\
HAS\_SUBSEQUENT\_VERSION & 79 \\
HAS\_SIMULATION\_TYPE & 74 \\
USES\_PHYSICAL\_SCHEME & 73 \\
HAS\_SUB\_EXPERIMENT & 72 \\
PART\_OF\_ENSEMBLE & 53 \\
INHERITED\_FROM & 3 \\
\bottomrule
\end{tabular*}
\end{table*}

\clearpage
\onecolumn
\raggedbottom
\subsection*{Prompts}
\label{subsec:app_prompts}

Prompt titles below use the same agent names as Figure~\ref{fig:info_extraction} and Section~\ref{subsec:ie_pipeline}. The document-level prompts (field, keywords, problem, method, task, innovation, model, observation and metric) belong to the native document understanding step, and the ontology conformance and evidence validator agents use no prompt, since both are deterministic.
\par\vspace{8pt}
\label{subsubsec:triplet_extraction_prompt}
\noindent\fcolorbox{black}{gray!20}{%
\begin{minipage}{0.95\textwidth}
\textbf{Title:} \textit{Triplet and evidence extractor agent prompt}

\textbf{Input:} \textit{PDF in native format(not parsed), ontology, entity\_definitions, relation\_definitions}

\textbf{Output:} \textit{Structured format: List of source entity class, source entity instance, relation, target entity class, target entity instance}

\textbf{Role Description:} You are an expert in extracting information from scientific papers. Your job is to answer the user's query based on the pdf file you have access to.

\textbf{Task:} You are given a scientific paper (or paragraph from it) related to climate modeling or Earth system science. The KG is primarily composed of ESGF metadata. Your task is to extract (source, relation, target) triplets with type that match the ontology below. For each extracted triplet, provide all supporting paragraphs from the text and explicitly resolve any anaphoras (e.g., "it", "this model") into their full entity names.
The ontology is given below in the form of (source type, target type, relation):

\vspace{0.5em}

\textbf{Ontology:} \texttt{\{ontology\}}

\vspace{0.5em}

\textbf{Sample ontology:}

\texttt{('SourceComponent','Variable','PRODUCES\_VARIABLE')}

\texttt{('SourceComponent','Realm','APPLIES\_TO\_REALM')}

\texttt{('Source','PhysicalScheme','USES\_PHYSICAL\_SCHEME')}

\vspace{0.5em}

\textbf{Entity type definitions:} \texttt{\{entity\_definitions\}}

\vspace{0.5em}

\textbf{Sample definitions:}

\texttt{\textbf{Source:} A climate model or system used to generate data (e.g., GFDL-ESM4).} 

\texttt{\textbf{SourceComponent:} A sub-model (e.g., atmosphere, ocean, land) within a coupled climate model. It may also function as a standalone climate model when used independently (e.g., ACCESS-OM2 as the ocean component in ACCESS-CM2).}

\texttt{\textbf{Realm:} Component of the Earth system (e.g., atmosphere, ocean). }

\texttt{\textbf{Variable:} Scientific quantities output by models (e.g., temperature, precipitation). }

\texttt{\textbf{PhysicalScheme:}  A representation of a physical process within a climate model (e.g., cloud scheme).}

\vspace{0.5em}

\textbf{Relation definitions:} \texttt{\{relation\_defintions\}}

\vspace{0.5em}

\textbf{Sample definitions:}

\texttt{\textbf{PRODUCES\_VARIABLE:} Specifies that a Source or SourceComponent outputs a particular Variable.}

\texttt{\textbf{APPLIES\_TO\_REALM:} Indicates that a Source, SourceComponent, or Experiment is associated with a specific Realm (e.g., atmosphere, ocean, land surface).}

\texttt{\textbf{USES\_PHYSICAL\_SCHEME:} Indicates that a Source employs a particular PhysicalScheme (e.g., radiation scheme, convection scheme).}

\vspace{0.5em}

\textbf{ONLY PROVIDE THE TRIPLETS IF THEY ARE PRESENT IN THE PAPER.}

\end{minipage}%
}
\par\vspace{8pt}

\label{subsubsec:app_selection_scientific_publication_prompt}
\noindent\fcolorbox{black}{gray!20}{%
\begin{minipage}{0.95\textwidth}
\textbf{Title:} \textit{Candidate paper selection prompt (corpus collection)}

\textbf{Input:} \textit{Title and Abstract}

\textbf{Output:} \textit{\texttt{yes} or \texttt{no}}

\textbf{Task:} Based on the title and abstract of the paper, analyze whether the paper likely contains climate model simulation data or results. Answer with a simple "yes" or "no".

\vspace{0.5em}

\end{minipage}%
}
\par\vspace{8pt}

\label{subsubsec:embedding_fallback_prompt}
\noindent\fcolorbox{black}{gray!20}{%
\begin{minipage}{0.95\textwidth}
\textbf{Title:} \textit{Hybrid entity linking: LLM fallback prompt}

\textbf{Input:} \textit{key\_word\_phrase, entity\_type, instances (candidate instance list), paragraph\_context}

\textbf{Output:} \textit{Exactly one instance string from the list, or \texttt{no match} (no explanation)}

\textbf{Role Description:} You are an expert at entity linking under schema constraints. You must pick the most likely instance match from a provided candidate list.

\textbf{Task:} Given the following key word/phrase of an entity type extracted from paragraph context, output which of the instances it most likely refers to. If none of the instances match, output \texttt{no match}. Do not provide an explanation.

\vspace{0.5em}
\textbf{Key Word/Phrase:} \texttt{\{key\_word\_phrase\}}

\textbf{Entity type:} \texttt{\{entity\_type\}}

\textbf{Instances:}
\texttt{\{instances\}}

\textbf{Paragraph context:}
\texttt{\{paragraph\}}

\vspace{0.5em}
\textbf{OUTPUT RULES:} Output \textbf{only} one of the given instances \textbf{exactly as written}, or \texttt{no match}. No punctuation, no extra text.
\end{minipage}%
}
\par\vspace{8pt}

\label{subsubsec:general_feature_extraction_prompt}
\noindent\fcolorbox{black}{gray!20}{%
\begin{minipage}{0.95\textwidth}
\textbf{Title:} \textit{Field extraction prompt}

\textbf{Input:} \textit{PDF in native format (not parsed)}

\textbf{Output:} \textit{One short phrase describing the research field (no extra text)}

\textbf{Role Description:} You are an expert at identifying the primary research field of scientific papers from their content.

\textbf{Task:} Summarize the research field of the paper in \textbf{one phrase}. 
Example: \textit{climate change}

\vspace{0.5em}

\end{minipage}%
}
\par\vspace{8pt}

\noindent\fcolorbox{black}{gray!20}{%
\begin{minipage}{0.95\textwidth}
\textbf{Title:} \textit{Keywords extraction prompt}

\textbf{Input:} \textit{PDF in native format (not parsed)}

\textbf{Output:} \textit{3--5 keywords separated by commas (single line, no bullet points)}

\textbf{Role Description:} You are an expert at extracting concise topical keywords that summarize a scientific paper.

\textbf{Task:} Extract \textbf{3--5 keywords} summarizing the content, topic, and field of the paper. Each keyword must be separated by a comma.
Example: \textit{climate change, global warming, greenhouse gases}

\vspace{0.5em}

\end{minipage}%
}
\par\vspace{8pt}

\noindent\fcolorbox{black}{gray!20}{%
\begin{minipage}{0.95\textwidth}
\textbf{Title:} \textit{Problem extraction prompt}

\textbf{Input:} \textit{PDF in native format (not parsed)}

\textbf{Output:} \textit{One sentence (max 50 words)}

\textbf{Role Description:} You are an expert at identifying the main problem statement a scientific paper addresses.

\textbf{Task:} Describe the problem the paper focuses on solving (e.g., shortcomings of existing methods, challenges in the task, factors affecting performance). Summarize in \textbf{one sentence (max 50 words)}.
Example: \textit{Existing climate models struggle to accurately simulate extreme precipitation events due to coarse spatial resolution and inadequate representation of convective processes.}

\vspace{0.5em}

\end{minipage}%
}
\par\vspace{8pt}

\noindent\fcolorbox{black}{gray!20}{%
\begin{minipage}{0.95\textwidth}
\textbf{Title:} \textit{Method extraction prompt}

\textbf{Input:} \textit{PDF in native format (not parsed)}

\textbf{Output:} \textit{One sentence (max 50 words)}

\textbf{Role Description:} You are an expert at summarizing proposed methods from scientific papers.

\textbf{Task:} Summarize the method proposed in the paper to solve the problem. Summarize in \textbf{one sentence (max 50 words)}.
Example: \textit{We propose a bias-correction method using quantile mapping to improve the accuracy of downscaled temperature projections from regional climate models.}

\vspace{0.5em}

\end{minipage}%
}
\par\vspace{8pt}

\noindent\fcolorbox{black}{gray!20}{%
\begin{minipage}{0.95\textwidth}
\textbf{Title:} \textit{Task extraction prompt}

\textbf{Input:} \textit{PDF in native format (not parsed)}

\textbf{Output:} \textit{One sentence (max 50 words)}

\textbf{Role Description:} You are an expert at identifying the main task a scientific paper focuses on.

\textbf{Task:} Summarize the task the paper focuses on solving. Summarize in \textbf{one sentence (max 50 words)}.
Example: \textit{The task is to project future temperature and precipitation patterns over South Asia using downscaled CMIP6 climate model outputs.}

\vspace{0.5em}

\end{minipage}%
}
\par\vspace{8pt}

\noindent\fcolorbox{black}{gray!20}{%
\begin{minipage}{0.95\textwidth}
\textbf{Title:} \textit{Innovation / contribution extraction prompt}

\textbf{Input:} \textit{PDF in native format (not parsed)}

\textbf{Output:} \textit{One sentence (max 50 words)}

\textbf{Role Description:} You are an expert at identifying the key innovation or contribution of a scientific paper.

\textbf{Task:} Describe the innovation or contribution of the paper in \textbf{one sentence (max 50 words)}.
Example: \textit{This paper introduces a novel bias-correction technique that significantly improves the accuracy of regional climate model projections for extreme precipitation events in South Asia.}

\vspace{0.5em}

\end{minipage}%
}
\par\vspace{8pt}

\label{subsubsec:paper_screening_prompt}
\noindent\fcolorbox{black}{gray!20}{%
\begin{minipage}{0.95\textwidth}
\textbf{Title:} \textit{Paper screening agent prompt (observation + metric)}

\textbf{Input:} \textit{PDF in native format (not parsed)}

\textbf{Output:} \textit{\texttt{yes} or \texttt{no} (single token)}

\textbf{Role Description:} You are an expert at screening climate-science papers for evaluation content.

\textbf{Task:} Does this paper have \textbf{(1) an observational dataset} which the model produces or is evaluated upon, \textbf{and} \textbf{(2) at least one quantitative evaluation metric}? 
Answer with \texttt{yes} only if it fulfills \textbf{both} conditions, otherwise answer with \texttt{no}.
Example: \texttt{yes} or \texttt{no}

\vspace{0.5em}

\end{minipage}%
}
\par\vspace{8pt}

\label{subsubsec:validation_agent_prompt}
\noindent\fcolorbox{black}{gray!20}{%
\begin{minipage}{0.95\textwidth}
\textbf{Title:} \textit{Evaluator prompt ($\TripletCorrectness$ judge)}

\textbf{Input:} \textit{(Source, Relation, Target) triplet and evidence}

\textbf{Output:} \textit{\texttt{valid} or \texttt{invalid}}

\textbf{Task:} You are given a paragraph related to climate modeling or Earth system science, and a (source instance, relation, target instance) triplet extracted from it. The KG is primarily composed of ESGF metadata. Your task is to verify if the triplet supports the paragraph from the paper. Assume anaphoras are there in the paragraph. Use your own knowledge to insert the source or target to the context of the paragraph.  Answer with 'valid' or 'invalid'. Do not provide an explanation.

\vspace{0.5em}

\textbf{Extracted Triplet:}

\textbf{Source:} \texttt{\{source\}}

\textbf{Relation:} \texttt{\{relation\}}

\textbf{Target:} \texttt{\{target\}}

\textbf{Paragraph context:} \texttt{\{paragraph\}}

\vspace{0.5em}

\end{minipage}%
}
\par\vspace{8pt}

\label{subsubsec:domain_specific_feature_extraction_prompt}
\noindent\fcolorbox{black}{gray!20}{%
\begin{minipage}{0.95\textwidth}
\textbf{Title:} \textit{Model extraction prompt}

\textbf{Input:} \textit{PDF in native format (not parsed)}

\textbf{Output:} \textit{Comma-separated list of climate model abbreviated names, or \texttt{no}}

\textbf{Role Description:} You are an expert at identifying climate model names mentioned in scientific papers.

\textbf{Task:} Name the climate model(s) this paper talks about. If nothing is mentioned, answer \texttt{no}. 
If present, each model name must be a \textbf{single abbreviated form} (no expansions in parentheses).
Correct example: \texttt{ACCESS-CM2}. 
Wrong example: \texttt{ACCESS-CM2 (Australian Community Climate and Earth System Simulator Climate Model Version 2).}

\vspace{0.5em}

\end{minipage}%
}
\par\vspace{8pt}

\noindent\fcolorbox{black}{gray!20}{%
\begin{minipage}{0.95\textwidth}
\textbf{Title:} \textit{Observation and metric extraction prompt (table-only)}

\textbf{Input:} \textit{PDF in native format (not parsed)}

\textbf{Output:} \textit{Structured list of entries: Model, ObservationalDataset, Metric, ResultValue(s)}

\textbf{Role Description:} You are an expert at extracting evaluation details from scientific paper tables with high precision.

\textbf{Task:} Extract the \textbf{observational dataset}, \textbf{evaluation metrics}, and \textbf{results} for \textbf{each climate model} \textbf{only from the table(s) in the paper}. 
\begin{itemize}
    \item Extract \textbf{only quantitative} metric values (numbers, percentages, error values, correlations, etc.), not descriptive statements.
    \item If a required field is missing in the tables, use \texttt{no}.
\end{itemize}

\vspace{0.5em}
\textbf{ONLY USE TABLE CONTENT. DO NOT INFER FROM CAPTIONS OR BODY TEXT.}
\end{minipage}%
}
\par\vspace{8pt}

\noindent\fcolorbox{black}{gray!20}{%
\begin{minipage}{0.95\textwidth}
\textbf{Title:} \textit{Validator agent prompt}

\textbf{Input:} \textit{Paragraph from a scientific paper and a candidate (source, source type, relation, target, target type)}

\textbf{Output:}  \texttt{valid} \textit{or} \texttt{invalid}

\textbf{Role Description:} You are an expert in climate modeling and Earth system science, with strong familiarity with ESGF metadata and related controlled vocabularies.

\textbf{Task:} Determine whether the given paragraph reasonably supports the extracted
(\textbf{source instance}, \textbf{relation}, \textbf{target instance}) triplet.
Each instance is accompanied by its semantic type (e.g., Source, Experiment, Variable, Institution, Region).

\begin{itemize}
    \item The source or target may not be mentioned verbatim but may be inferred from context.
    \item Domain knowledge may be used to resolve abbreviations or implicit mentions.
    \item The paragraph may contain anaphora or implicit references.
\end{itemize}

The triplet should be considered \textbf{valid} if the relationship is explicitly stated or clearly implied by the paragraph.
It should be considered \textbf{invalid} only if the relationship is contradicted or unsupported by the text.

\vspace{0.5em}
\textbf{Respond with \emph{only} one of the following labels:}
\begin{itemize}
    \item \texttt{valid}
    \item \texttt{invalid}
\end{itemize}

\textbf{Do not provide explanations or additional text.}

\vspace{0.5em}
\textbf{Extracted Triplet:}
\begin{itemize}
    \item \textbf{Source:} \{source\}
    \item \textbf{Source Type:} \{source\_type\}
    \item \textbf{Relation:} \{relation\}
    \item \textbf{Target:} \{target\}
    \item \textbf{Target Type:} \{target\_type\}
\end{itemize}

\textbf{Paragraph context:}

\noindent\textit{\{paragraph\}}
\end{minipage}%
}
\par\vspace{8pt}

\end{document}